\documentclass[letterpaper,twocolumn,10pt,hyphens]{article}
\usepackage{usenix-2020-09}

\usepackage{tikz}
\usepackage{amsmath}
\usepackage{enumitem}
\usepackage{filecontents}
\usepackage{xspace}
\usepackage{float}
\usepackage{booktabs}
\usepackage{graphicx}
\usepackage{subcaption}
\usepackage{amsmath}
\usepackage{balance}

\usepackage{amssymb}
\usepackage{makecell}
\usepackage{array}
\usepackage[ruled,linesnumbered]{algorithm2e}
\usepackage{listings}
\usepackage{graphicx}
\usepackage{subcaption}
\usepackage{multirow}
\usepackage[table]{xcolor}
\usepackage{hyperref}          
\usepackage{breakurl}          

\newcommand{\system}{RLinf\xspace} 
\newcommand{\para}[1]{{\vspace{5pt} \bf \noindent #1 \hspace{5pt}}}

\newcommand{\squishlist}{
	\begin{list}{$\bullet$}
		{ \setlength{\itemsep}{0pt}      \setlength{\parsep}{3pt}
			\setlength{\topsep}{3pt}       \setlength{\partopsep}{0pt}
			\setlength{\leftmargin}{3.5mm} \setlength{\labelwidth}{1em}
			\setlength{\labelsep}{0.5em} } }
	\newcommand{\squishend}{
\end{list}  }

\definecolor{codekw1}{rgb}{0,0.5,0}
\definecolor{codekw2}{rgb}{0.69,0,0.25}
\definecolor{codekw3}{rgb}{0.75,0,0}
\definecolor{codekw4}{rgb}{0.15, 0.5, 0.6}
\definecolor{at}{rgb}{0.64,0.29,0.64}
\definecolor{codeln}{rgb}{0.5,0.5,0.5}
\definecolor{codestr}{rgb}{0,0.6,0}

\lstdefinelanguage{myC}{
    morekeywords=[1]{interface},
    morekeywords=[2]{TypedTile, tTile},
    morekeywords=[3]{if, else, for, in, return, def, with, while},
    morekeywords=[4]{__global__, __device__, void, @overload, @property},
    morekeywords=[5]{size\_t, vector, set, Config, struct, TileShape, TileType, TensorExpr, class}, 
    morekeywords=[6]{get\_input, tTile-Operator},
    sensitive=true,
    morecomment=[l]{\#},
    morecomment=[l]{//},
    morecomment=[n]{/*}{*/},
    morestring=[b]",
    morestring=[b]',
    morestring=[b]"""
}

\begin{document}

\date{}
\title{\Large \bf \system: Flexible and Efficient Large-scale Reinforcement Learning via Macro-to-Micro Flow Transformation}

%
\author{
	{\rm Chao Yu$^{12}$, 
	     Yuanqing Wang$^{34}$,
	     Zhen Guo$^{3}$,
	     Hao Lin$^{3}$,
	     Si Xu$^{3}$,
		 Hongzhi Zang$^{1}$,
          Quanlu Zhang$^{3}$,
    }
    \vspace{-1mm}
	\and
	{\rm 
        Yongji Wu$^{5}$,
        Chunyang Zhu$^{3}$,
	    Junhao Hu$^{3}$,
        Zixiao Huang$^{1}$,
        Mingjie Wei$^{2}$,
        Yuqing Xie$^{1}$,
        Ke Yang$^{2}$,
    }
    \vspace{-1mm}
   \and
	{\rm 
         Bo Dai$^{6}$,
         Zhexuan Xu$^{1}$,
         Jiakun Du$^{1}$,
         Xiangyuan Wang$^{4}$,
         Xu Fu$^{3}$,
         Letong Shi$^{3}$,
         Zhihao Liu$^{2}$,
         Kang Chen$^{42}$,
    }
    \vspace{-1mm}
    \and
    {\rm
         Weilin Liu$^{3}$,  
         Gang Liu$^{1}$,
        Boxun Li$^{3}$,
        Jianlei Yang$^{6}$,
         Zhi Yang$^{4}$,
         Guohao Dai$^{73}$,
         Yu Wang$^{1*}$
   }
    \and 
    {\rm
    $^{1}$Tsinghua University \hspace{0.3cm}
	$^{2}$Zhongguancun Academy \hspace{0.3cm}
	$^{3}$Infinigence AI  \hspace{0.3cm}
 }
 \vspace{-1mm}
    \and
    {\rm
    $^{4}$Peking University \hspace{0.3cm}
	$^{5}$UC Berkeley \hspace{0.3cm}
	$^{6}$Beihang University \hspace{0.3cm}
    $^{7}$Shanghai Jiaotong University
 }
 \and
    {\rm
    $^{*}$Corresponding Author: \url{yu-wang@tsinghua.edu.cn}
 }
 \and
    {\rm
    GitHub Repo: \url{https://github.com/RLinf/RLinf}
 }
}

\maketitle

\begin{abstract}
Reinforcement learning (RL) has demonstrated immense potential in advancing artificial general intelligence, agentic intelligence, and embodied intelligence.
However, the inherent heterogeneity and dynamicity of RL workflows often lead to low hardware utilization and slow training on existing systems.
In this paper, we present \system, a high-performance RL training system based on our key observation that the major roadblock to efficient RL training lies in \textit{system flexibility}.
To maximize flexibility and efficiency, 
    \system is built atop a novel RL system design paradigm called \textit{macro-to-micro flow transformation} (M2Flow),
    which automatically breaks down high-level, easy-to-compose RL workflows at both the temporal and spatial dimensions, 
    and recomposes them into optimized execution flows.
Supported by \system worker's adaptive communication capability,
    we devise \textit{context switching} and \textit{elastic pipelining} to realize M2Flow transformation,
    and a profiling-guided scheduling policy to generate optimal execution plans.
Extensive evaluations on both reasoning RL and embodied RL tasks demonstrate that \system consistently outperforms state-of-the-art systems, achieving 1.07$\times$$\sim$2.43$\times$ speedup in end-to-end training throughput.
\end{abstract}

\section{Introduction}
The rapid progress of large language models (LLMs) has reached a point where further scaling model alone yields diminishing returns. 
To push intelligence beyond pretraining, reinforcement learning (RL) has emerged as a crucial paradigm. 
Recent advances such as RLHF~\cite{rlhf,deeprl}, GRPO~\cite{grpo}, and RL for embodied agents~\cite{kim24openvla,simplevla} and Deep Research~\cite{nakano2021webgpt,zheng2025deepresearcher} all rely on RL to align LLMs with human preferences, improve reasoning, and enable autonomous interaction with complex environments. 
OpenAI and others predict that RL workloads will soon consume more computational resources than LLM pretraining~\cite{learningtoreasoning}, 
    making RL training efficiency the most critical system concern.

However, efficient RL training for various scenarios such as reasoning, agentic and embodiment at the scale of modern large models is challenging,
    which combines highly heterogeneous components with diverse workload characteristics and resource demands,
    such as LLM generation, inference and training, reward models, critic models, agent tooling and embodied environment simulators.
For instance,
    LLM training consumes more accelerator (e.g., GPU) memory than LLM generation and inference (prefill-only generation) to maintain gradients and optimizer states,
    while LLM generation shows high dynamicity in response lengths, 
    leading to low accelerator utilization. 
Moreover, components like LLM training support diverse parallelization strategies (e.g., data, tensor, pipeline parallelism), 
    whereas others scale only via instance replication and may yield computation workloads distinct from common tensor computation in LLM, e.g., 
    embodied simulators~\cite{duan2022survey,taomaniskill3} that require CPU for physics simulation and GPU graphics pipeline for 3D rendering. 

Single execution mode of existing RL training systems fails to capture this diversity,
    leading to suboptimal efficiency.
Collocated execution, where components sequentially occupy accelerators~\cite{verl}, 
    suffers from the long-tail problem due to varying generation lengths, 
    leaving accelerators idle. 
Disaggregated pipelining, where components run concurrently on separate accelerators with pipelining~\cite{areal}, 
    mitigates the long-tail issue but introduces memory and computation imbalance (\S\ref{sec:background:inefficiencies}). 
Neither mode is universally optimal. 
Many RL workloads demand hybrid scheduling of the components to maximize efficiency, 
    i.e., mixing collocation and pipelining in a more flexible way. 
However, supporting such flexible execution modes for a single programmed workflow is a significant challenge,
    as they often require different program structures and communication patterns.
Also, identifying the right scheduling for a given workflow usually requires considerable manual tunning.

In this paper, we present \system, an RL training system that maximizes system flexibility to achieve efficient execution of a logically programmed RL workflow.
At its core is a new paradigm called \textit{macro-to-micro flow transformation} (M2Flow), 
    i.e., macro logical flow with micro execution flow,
    which decouples the logical programming of RL workflows from their physical execution planning.
With M2Flow, developers program RL workflows imperatively, using a natural programming interface to define how components communicate and synchronize at a coarse granularity. 
\system then automatically transforms this logical flow into a fine-grained execution plan tailored to the workload and hardware at both spatial and temporal dimensions.
By decoupling program semantics from execution modes, 
    M2Flow lets developers preserve clean, intuitive workflows while the system explores a vast scheduling space, 
    including temporal multiplexing, spatial pipelining, and hybrid scheduling.

\system achieves this through three key mechanisms. 
First, a \textit{worker} abstraction that encapsulates each RL component for flexible placement,
    and built-in adaptive communication that allows direct and efficient communication between components regardless of worker and data placement.
Second, elastic pipelining and automatic context switching that enable M2Flow transformation and expand the scheduling space, 
    achieving pipeline granularity tunning and temporal accelerator multiplexing without modifying the logical workflow. 
Third, a profiling-guided scheduling policy that automatically selects efficient execution modes, balancing utilization across heterogeneous components. 
Together, these capabilities deliver both high flexibility, efficiency and programmability.

We implement \system using Ray for cluster management and worker process launch on remote nodes.
Apart from the core mechanisms, 
    \system also provides rich support for common RL components, algorithms, and models to facilitate RL workflow programming (\S\ref{sec:implementation}).
We extensively evaluate \system on both reasoning and embodied RL training across diverse models (e.g., Qwen2.5-1.5/7/32B~\cite{qwen2025qwen25technicalreport}, Qwen3~\cite{yang2025qwen3}, OpenVLA~\cite{kim24openvla}, OpenVLA-OFT~\cite{kim2025fine}) and varying scales. 
Our evaluation shows that \system improves end-to-end throughput (i.e., tokens per second) by up to 1.7$\times$ compared to state-of-the-art RL training systems in reasoning RL, 
    and by up to 2.43$\times$ in embodied RL. 
We have open-sourced the full codebase of \system to accelerate RL innovations in LLM era.

This paper makes the following contributions:
\squishlist
\item Analysis of representative RL algorithms and scenarios, identifying characteristics of modern RL workflows and highlighting inefficiencies in current systems.

\item A novel RL system paradigm M2Flow that decouples logical workflow programming from execution planning, enabling intuitive programming with flexible execution.

\item A cohesive set of mechanisms, i.e., worker abstraction, elastic pipelining, context switching, adaptive communication, and a profiling-guided scheduler, that jointly realize M2Flow and enable efficient RL training.

\item Comprehensive evaluations of \system across various RL workloads show that \system significantly improves efficiency and flexibility compared to existing approaches.
\squishend

\section{Background and Motivation}
\subsection{RL Workflows in LLM Era}
\label{sec:background:rl_workflows}

\begin{figure}[t]
    \centering
    \includegraphics[width=0.95\linewidth]{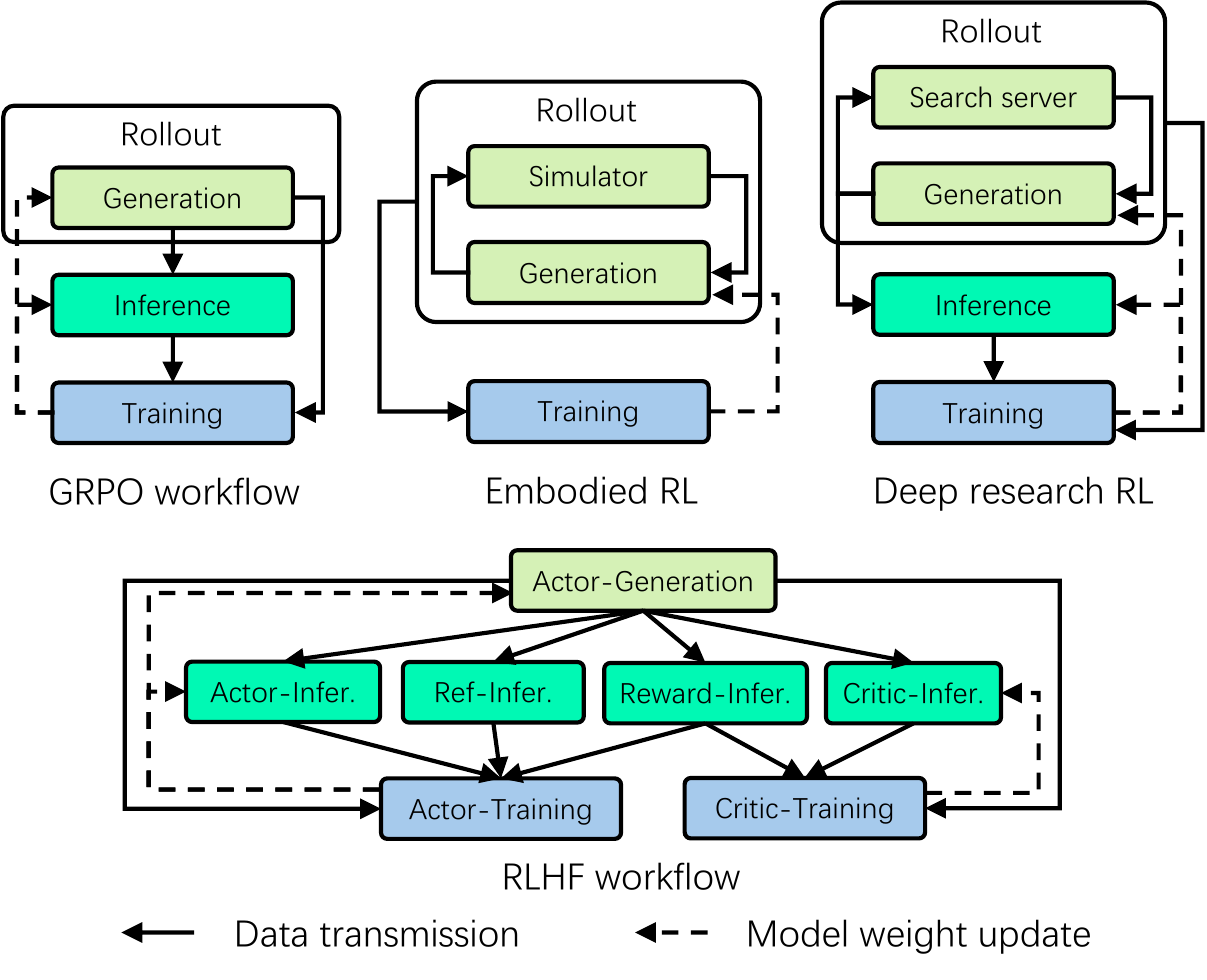}
    \vspace{-10pt}
    \caption{Diverse RL workflows in various scenarios.}
    \label{fig:ml_workflows}
    \vspace{-15pt}
\end{figure}

\para{Various RL Algorithms and Scenarios.}
With the slowdown of scaling gains in large language models, 
    reinforcement learning has become increasingly important for advancing LLM intelligence. 
Unlike traditional RL, RL in the LLM era often involves multiple LLMs in the loop. 
Given the scale of modern models (tens to hundreds of billions of parameters), 
    fitting RL training into available accelerators (e.g., GPUs) is already challenging. 
\autoref{fig:ml_workflows} illustrates four representative RL workflows across different scenarios and algorithms.

The simplest is GRPO~\cite{grpo}, an RL algorithm designed to reduce reliance on reward models. 
It involves a single LLM that generates multiple responses, e.g., 8, for a query (i.e., \texttt{Generation}), 
    computes logarithmic probabilities for these responses (i.e., \texttt{Inference}), 
    and uses the results as training data to update the same model (i.e., \texttt{Training}). 
The updated weights are then synchronized back for inference and generation, completing one training iteration.

In contrast, the RLHF~\cite{rlhf} workflow adopts PPO~\cite{ppo}, resulting in a more complex design involving four LLMs. The actor model serves as the core policy, generating responses to queries. The reference model remains fixed to constrain the actor from drifting too far from its initialization. The reward model assigns scalar rewards to generated responses, while the critic model estimates expected rewards to stabilize training. Actor and critic are trainable, whereas reference and reward models are frozen. These components interact closely, as shown in the figure.

Beyond algorithms, RL workflows also vary by application scenario. In embodied intelligence~\cite{kim24openvla,simplevla}, RL relies on simulators that simulates the physical world. 
An LLM interacts with the simulator by generating actions and receiving feedback, producing trajectories that serve as training data. 
Similarly, in Deep Research~\cite{nakano2021webgpt,zheng2025deepresearcher}, RL improves model performance through interaction with a search server that retrieves online information. 
The resulting rollout results are fed into training, while inference follows the GRPO workflow.

\para{Characteristics of RL Workflows.}
RL workflows consist of heterogeneous components with distinct demands on GPU memory, computation cores, accelerator types, and parallelization strategies. 
For example, training requires substantially more GPU memory than inference to maintain gradients and optimizer states. 
Unlike training, generation often under-utilizes GPU cores, as its matrix and vector multiplications are bottlenecked by memory bandwidth. 
Some components (e.g., simulator) run on CPUs, or use GPUs for non-tensor computations (e.g., 3D rendering).
Parallelization also differs significantly, 
    e.g., LLM training exploits data, tensor, and pipeline parallelism, whereas simulators typically scale only through multiple instances. 
Maximizing overall utilization across such heterogeneous components is a great challenge.

Further, RL workflows exhibit complex dependencies, primarily through data flow and weight updates. 
Data flow can occur at different granularity, e.g., 
    per response between generation and inference, 
    or at least a micro-batch of responses between generation and training. 
Some workflows even introduce cyclic data flows, 
    such as in embodied RL and Deep Research (\autoref{fig:ml_workflows}), 
    which further complicates coordination. 
Weight updates, in contrast, act as barriers that synchronize generation and training. 
These complex dependencies require more fine-grained scheduling.

\subsection{Inefficiencies in Diverse RL Workflows}\label{sec:background:inefficiencies}
We analyzed different RL workflows to identify the source of inefficiencies as follows.

\begin{figure}[t]
    \centering
    \begin{subfigure}[b]{0.49\linewidth}
        \includegraphics[width=\linewidth]{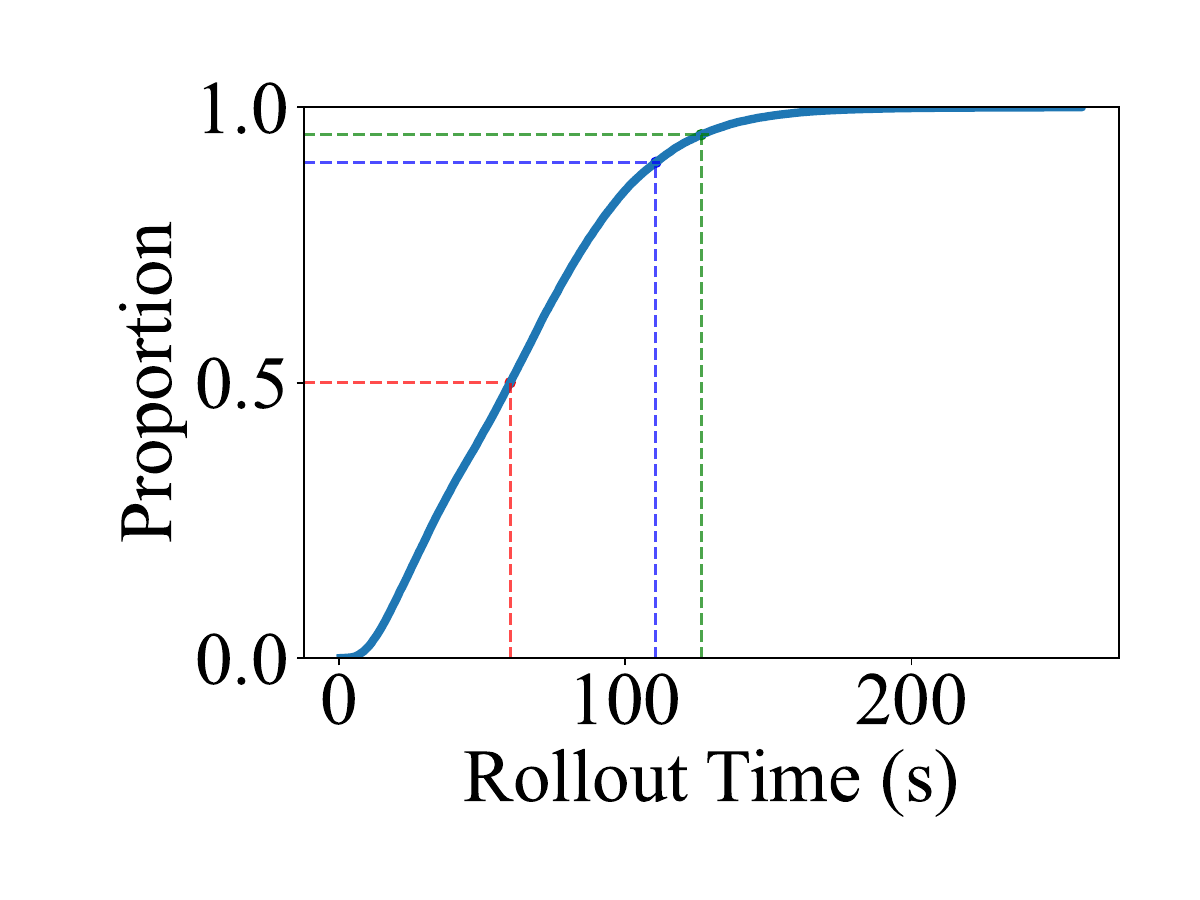}
        \vspace{-15pt}
        \caption{CDF of response time.}
        \label{fig:rollout_response_cdf}
    \end{subfigure}
    \begin{subfigure}[b]{0.49\linewidth}
        \includegraphics[width=\linewidth]{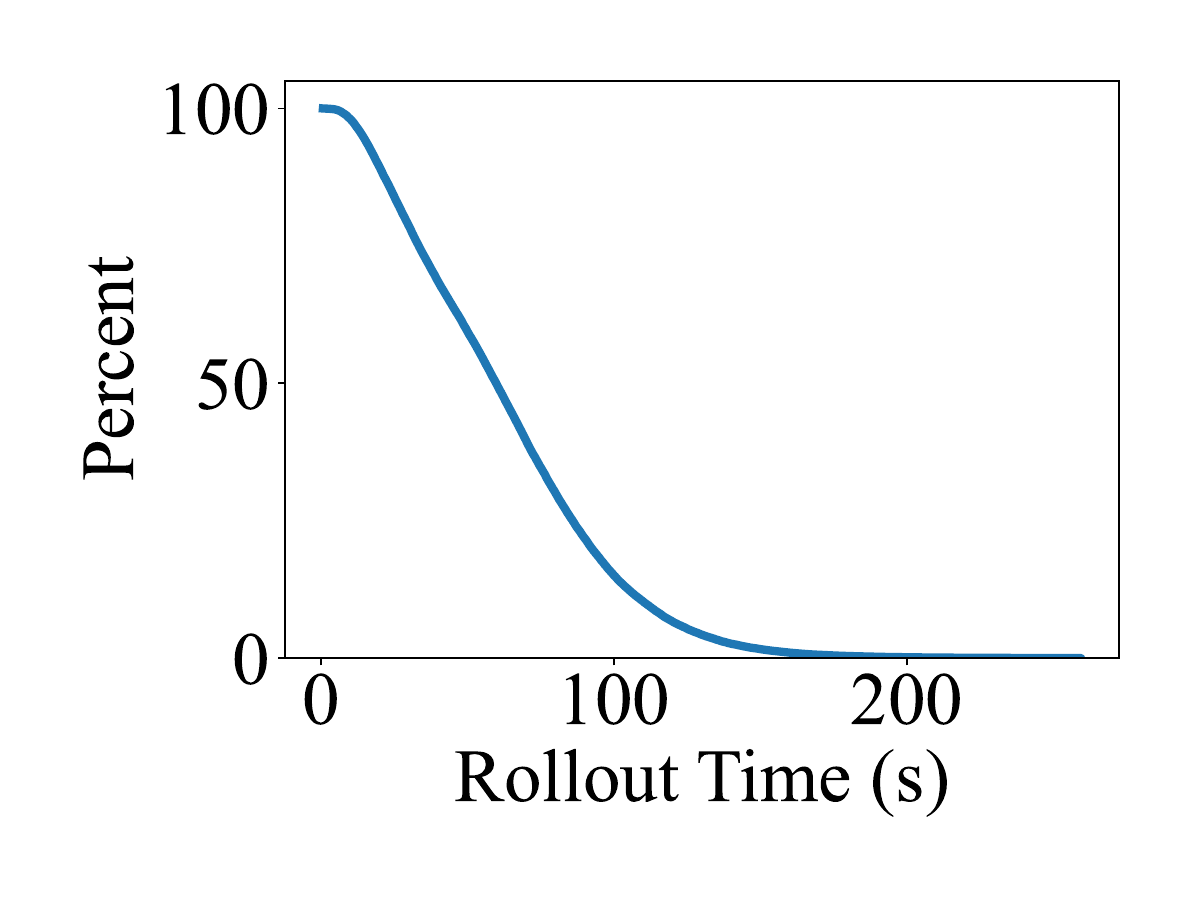}
        \vspace{-15pt}
        \caption{Unfinished responses.}
        \label{fig:roll_response_time}
    \end{subfigure}
    \caption{The distribution of response lengths and the number of unfinished responses over time in the generation phase of a math RL experiment.}
    \vspace{-15pt}
    \label{fig:rollout_responses}
\end{figure}

\para{Dynamicity in Rollout Wastes Computation.}
The rollout phase is inherently dynamic. Lengths fluctuate across responses of the same query, and even more so across different queries. Embodied tasks such as grasping can take different numbers of steps. In deep research, generating a report may involve varying numbers of search interactions. Since rollouts are executed in batches, these variations create a long-tail problem where a few slow queries block the entire phase from proceeding to inference or training. The problem commonly exists in the collocated mode (e.g., veRL~\cite{verl}), where generation, inference, and training sequentially share GPUs by swapping between CPU and GPU memory. We conducted an experiment of a 7B math reasoning RL training~\cite{qwen2025qwen25technicalreport} on 8 nodes with 8 H100 GPUs each, as shown in \autoref{fig:rollout_responses}. The number of unfinished responses quickly shrinks to less than 5\%, where a very small set of long-tail responses stalls the generation, leaving many GPUs underutilized or idle. Scaling out with more GPUs worsens the problem as idle time grows.

Pipelining alleviates this by allocating fewer GPUs to generation and leaving the rest for inference and training. In this setup, inference and training start once partial samples are ready. However, pipelining introduces its own inefficiency since inference and training must wait for the first batch to be generated. Thus, neither approach is universally optimal, and supporting both collocated mode and pipelining within a single framework remains a significant challenge.

\begin{figure}[t]
    \centering
    \begin{subfigure}[b]{0.49\linewidth}
        \includegraphics[width=\linewidth]{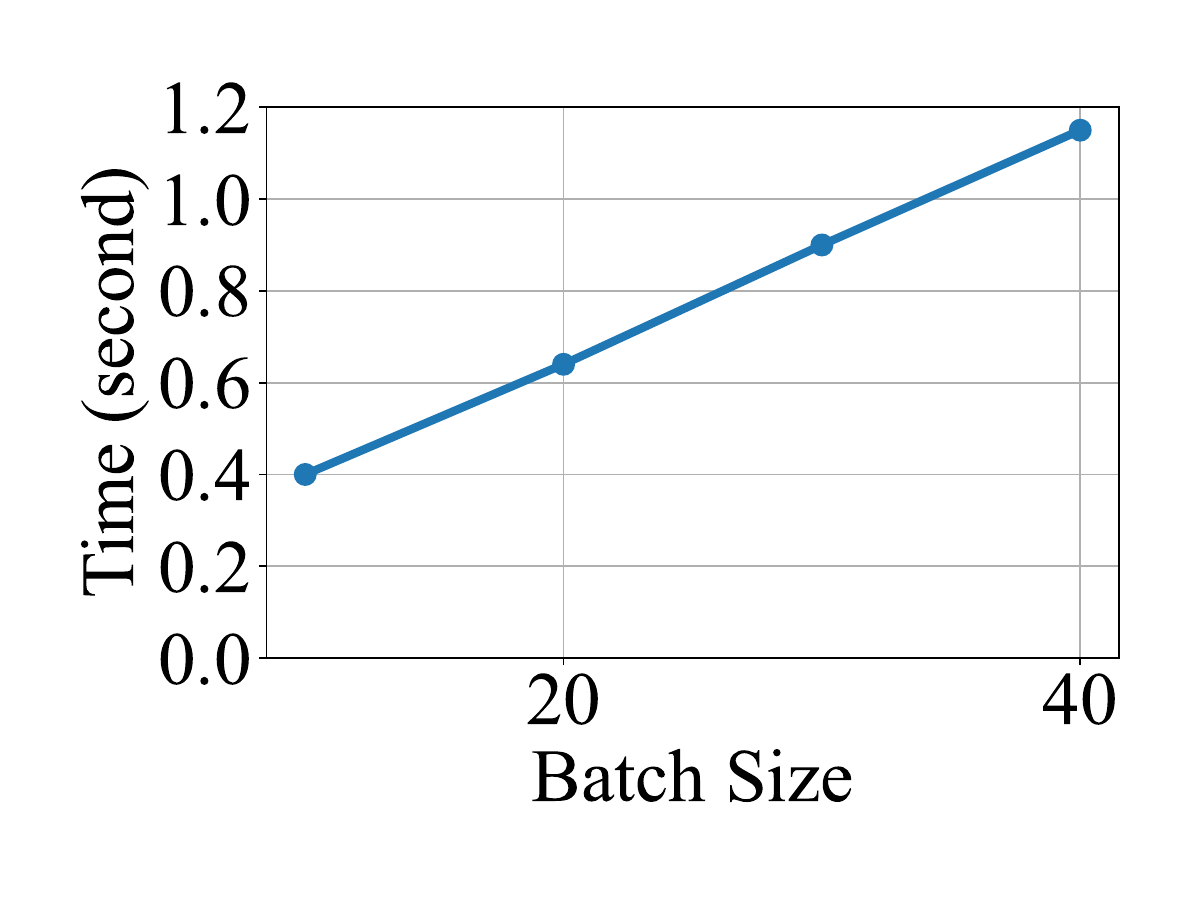}
        \vspace{-15pt}
        \caption{Generation time.}
        \label{fig:time_vs_batchsize}
    \end{subfigure}
    \begin{subfigure}[b]{0.49\linewidth}
        \includegraphics[width=\linewidth]{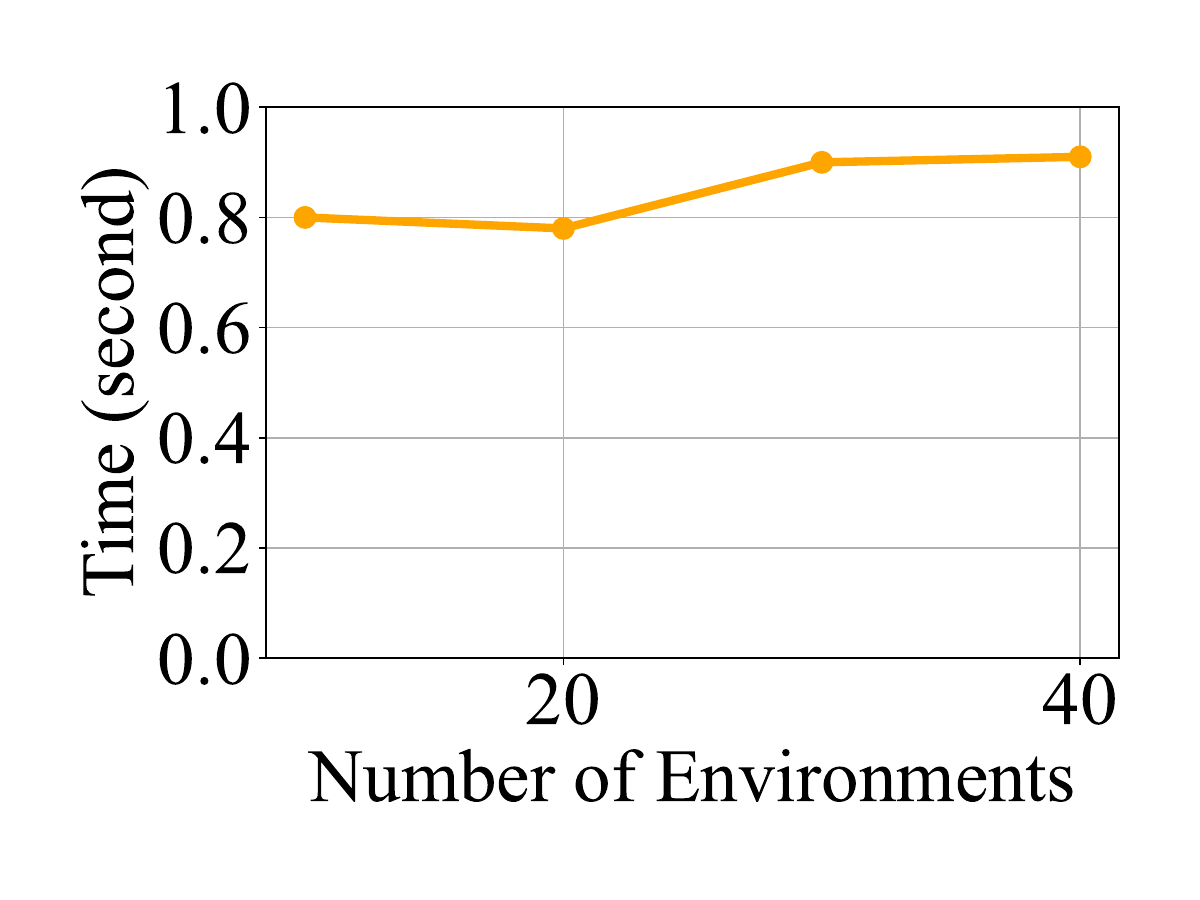}
        \vspace{-15pt}
        \caption{Simulator time.}
        \label{fig:time_vs_envs}
    \end{subfigure}
    \caption{The execution time of generation and simualtor with different batch sizes respectively, batch size in simulator is the number of environments.}
    \vspace{-15pt}
    \label{fig:embodied_gpu_utilizations}
\end{figure}

\para{Simple Execution Modes Cannot Fit Diverse Components.}
Collocated and pipelined modes are two extremes, i.e., all components on the same GPUs versus fully disaggregated GPUs. Some RL workloads, however, do not fit neatly into either mode. Their diverse component characteristics require more flexible orchestration.
Take embodied RL (\autoref{fig:ml_workflows}) as an example. \autoref{fig:embodied_gpu_utilizations} shows computation profiles of generation and simulator. The execution time of simulator increases slightly with the number of environments and its GPU utilization remains low (i.e., <24\%), while its memory usage grows linearly with the number of environments. In contrast, generation scales linearly in both runtime and memory with batch size, while keeping GPU cores highly utilized (i.e., >70\%). Training consumes more memory, but its execution time is only one-third as long as generation’s.

This profile rules out simple execution modes. The simulator should scale with as many parallel environments as possible to reduce its total runtime. However, this prevents collocation with generation due to memory contention. A better choice is running on disaggregated GPUs with pipelining for higher efficiency. Training, in contrast, would waste compute if fully disaggregated, so it should share GPUs. After rollout, simulator and generation are swapped to CPU, and training takes over the GPU. The result is a hybrid mode that combines collocation and pipelining to balance efficiency.

\para{Identifying Suitable Orchestration is Challenging.}
The orchestration of the hybrid mode depends on the analysis of the components. However, finding the most suitable orchestration for a given RL workflow is challenging, as the characteristics are diverse and the dependencies are complex. Manually enumerating options is tedious, time-consuming, and risks overlooking better choices. Moreover, no clear guidelines exist to identify the most suitable orchestration.

\subsection{Flexibility as a Key to Efficiency}
Maximizing computation efficiency for an RL workload requires flexible orchestration that aligns with component and workflow characteristics. However, adjusting the execution mode without changing the programmed workflow is challenging. Collocated and disaggregated pipelining modes differ significantly. Collocated mode operates at coarse-grained, phase-level execution. Each phase starts after the previous phase is complete. However, disaggregated pipelining runs at fine-grained, batch-level with precise timing to minimize pipeline bubbles. Mixing these modes further increases complexity. We advocate a system design that bridges this gap, enabling RL developers to maintain an intuitive, logically organized workflow while achieving high execution efficiency with flexible execution modes.

\section{\system Design}

\subsection{Overview}
\begin{figure}[t]
    \centering
    \includegraphics[width=0.9\linewidth]{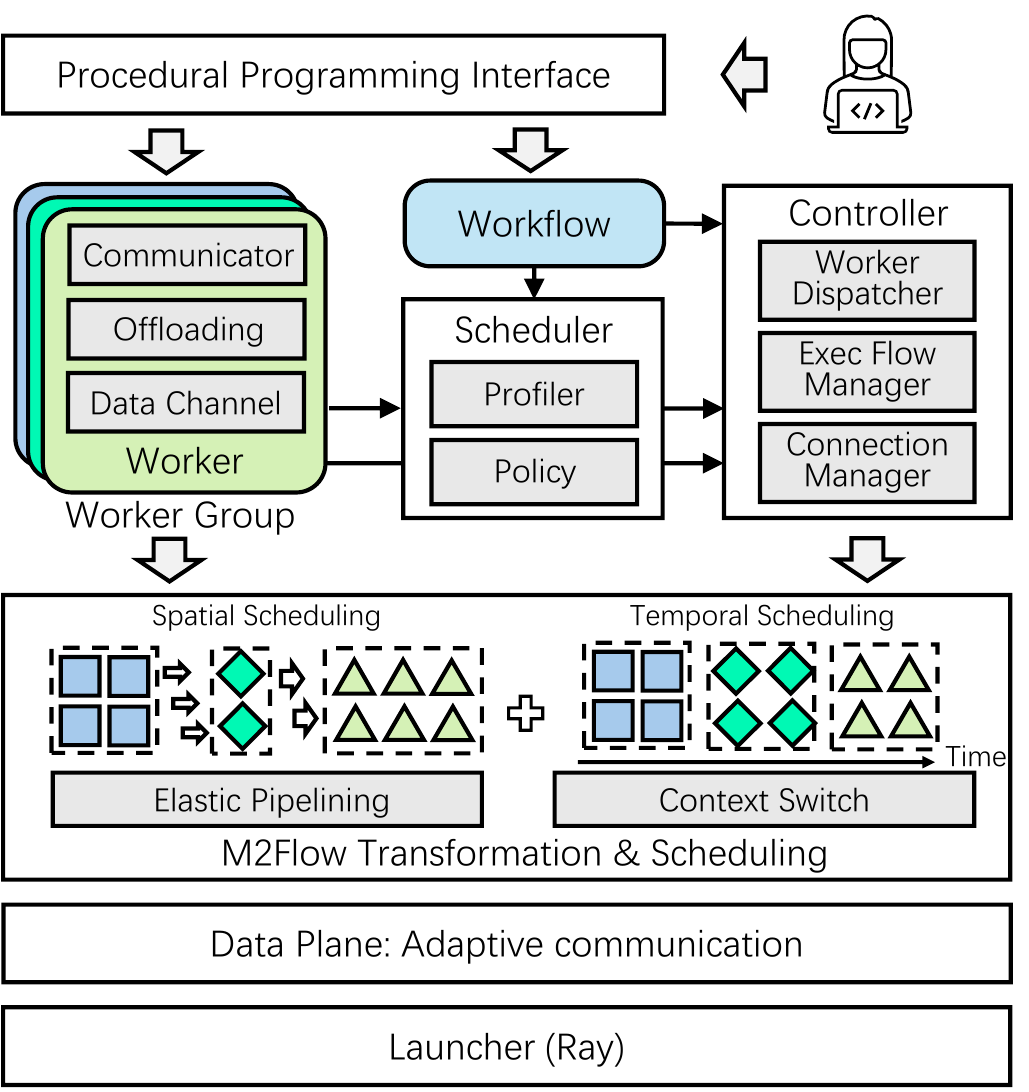}
    \caption{The architecture of \system.}
    \vspace{-15pt}
    \label{fig:arch}
\end{figure}

In pursuit of efficient, flexible, and intuitive RL systems,
  we propose a new design paradigm termed M2Flow, i.e., \textit{macro logical flow executed with micro execution flow}.
In this paradigm, developers program the complex RL workflow by imperatively specifying the logical communication flow among the RL components at a coarse granularity (macro logical flow),
  while the system automatically transforms the workflow into a fine-grained execution flow (micro execution flow).
Essentially,
  M2Flow decouples programmable code logic from the physical execution and scheduling of the individual RL components,
  so as to maximize the efficiency while minimizing the programming complexity.

\autoref{fig:arch} shows the architecture of \system that realizes M2Flow.
As illustrated,
  \system provides an easy-to-use procedural programming interface for users to construct RL workflows imperatively,
  which describe the data communication and interaction among the RL components.
RL components are then encapsulated as \textit{workers} in \system, 
  each implementing the main logic of this component.
Workers are equipped with communication functionality to freely communicate with each other,
  as well as resource offloading mechanism to enable temporal multiplexing of hardware resources.
This worker abstraction enables \system to retain substantial scheduling flexibility at both the spatial and temporal dimensions, while following the procedural workflow.
Spatial scheduling assigns workers to accelerators, temporal scheduling determines their execution periods, and spatio–temporal scheduling governs their pipelined execution granularity.
 
To produce a desirable micro execution flow across these scheduling dimensions,
  the core of the scheduler module is a profiling-guided scheduling policy,
  which utilizes runtime profiling of worker characteristics to search for the optimal execution mode of each worker.
Based on the determined execution mode, the Controller assigns workers to accelerators, manages inter-worker connections, and orchestrates the execution flow by dispatching function invocations to the corresponding workers.
To realize this, two mechanisms named elastic pipelining and context switch are devised to enable spatial and temporal orchestration of workers, respectively. 
Adaptive communication utilities such as point-to-point communication and data channel act as the data plane to support scalable worker interactions. 
The entire system leverages Ray~\cite{ray} to remotely launch and control workers.

\subsection{Workflow Construction Interface}
\begin{figure}[t]
\centering
\begin{subfigure}[t]{0.98\linewidth}
\begin{lstlisting}[numbers=none, xleftmargin=0em,framexleftmargin=0em,showlines=true]
class Worker:
  def send(self, obj, dst, async_op)
  def recv(self, src, async_op)
  @overload
  def onload(self)
  @overload
  def offload(self)

class RolloutWorker(Worker):
  def generate(self, in_channel, out_channel):
    with self.device_lock:
      batch_size = 0
      while batch_size < self.total_batch_size:
        batch = in_channel.get()
        res = self.model.gen(batch)
        out_channel.put(res)
        batch_size += batch.size
\end{lstlisting}
\vspace{-10pt}
\caption{A typical \system worker.}
\label{fig:programinterface_worker}
\end{subfigure}

\begin{subfigure}[t]{0.98\linewidth}
\begin{lstlisting}[numbers=none, , xleftmargin=0em,framexleftmargin=0em]
class RLWorkflowRunner:
  def __init__(self):
    cluster = Cluster(num_nodes=4, devices_per_node=8)
    self.rollout_group = RolloutWorker.launch(cluster)
    self.actor_group = ActorWorker.launch(cluster)
    self.data_ch = Channel.create("Data")
    self.rollout_ch = Channel.create("Rollout")
  def run(self):
    for batch in batch_iterator:
      self._update_rollout_weights()
      self.data_ch.put(batch)
      self.rollout_group.generate(
        in_channel=self.data_ch,
        out_channel=self.rollout_ch
      )
      self.actor_group.train(self.rollout_ch).wait()
\end{lstlisting}
\vspace{-10pt}
\caption{A workflow runner example.}
\label{fig:programinterface_runner}
\end{subfigure}

\caption{\system workflow programming interface.}
\vspace{-15pt}
\label{fig:programinterface}
\end{figure}


The design philosophy of \system is to maximize system flexibility to achieve high efficiency,
  which is also the guideline for \system's programming interface design.
To this end, unlike traditional graph-based declarative programming~\cite{abadi2016tensorflow} that sacrifices control flow flexibility, debuggability and transparency for optimization opportunity, 
  \system adopts a procedural programming paradigm that enables developers to flexibly express workflows imperatively. 
An example workflow based on this interface is shown in \autoref{fig:programinterface}.
As shown, an RL program based on \system consists of two parts: (1) worker programs that define the logic of each RL component (e.g., simulator, LLM generation, actor training), and (2) a workflow runner that orchestrates the overall workflow by invoking the workers' core functions and defining inter-worker interactions.

\autoref{fig:programinterface_worker} demonstrates a typical worker implementation atop \system.
The base \texttt{Worker} class provides communication primitives such as \texttt{send} and \texttt{recv} for inter-worker communication.
All workers inherit from the base class automatically gains the capability to communicate with other workers (\S\ref{sec:design_communication}),
  which is also the foundation for higher-level communication facilities like the data channel.
To manage limited device resources like GPU memory, 
  all workers are required to implement resource management functions (\texttt{onload} and \texttt{offload}) for acquiring and releasing the resources.

After implementing workers, developers can compose the overall RL workflow as in \autoref{fig:programinterface_runner}.
First, the runner launches workers on a cluster of nodes and devices in an SPMD manner.
The scheduler module decides the placement of each worker process before its launch,
  which can also be manually specified if desired.
All processes of the same worker are collectively managed via the \texttt{WorkerGroup} abstraction of \system (e.g., \texttt{rollout\_group}),
  which automatically assumes the public functions defined in the worker class and dispatches them to all (or a selective portion) of the worker processes if invoked.
The functions of a \texttt{WorkerGroup} is inherently asynchronous and returns a result handle, 
  whose \texttt{wait} primitive provides synchronization barriers, enabling the computations on certain granularity of data.
For example, in GRPO training, 
  rollout can proceed per query, 
  but normalization must aggregate all responses for a query,
  which pauses the pipeline at this step until normalization completes.
The core facility for connecting the data flow among distributed worker groups is the data channel (detailed in \S\ref{sec:design_communication}), which decouples the control and data flows of dependent components, enabling highly flexible programming while exposing a broad optimization space, as shown below.

\subsection{M2Flow Transformation}
\begin{figure}[t]
    \centering
    \includegraphics[width=1.0\linewidth]{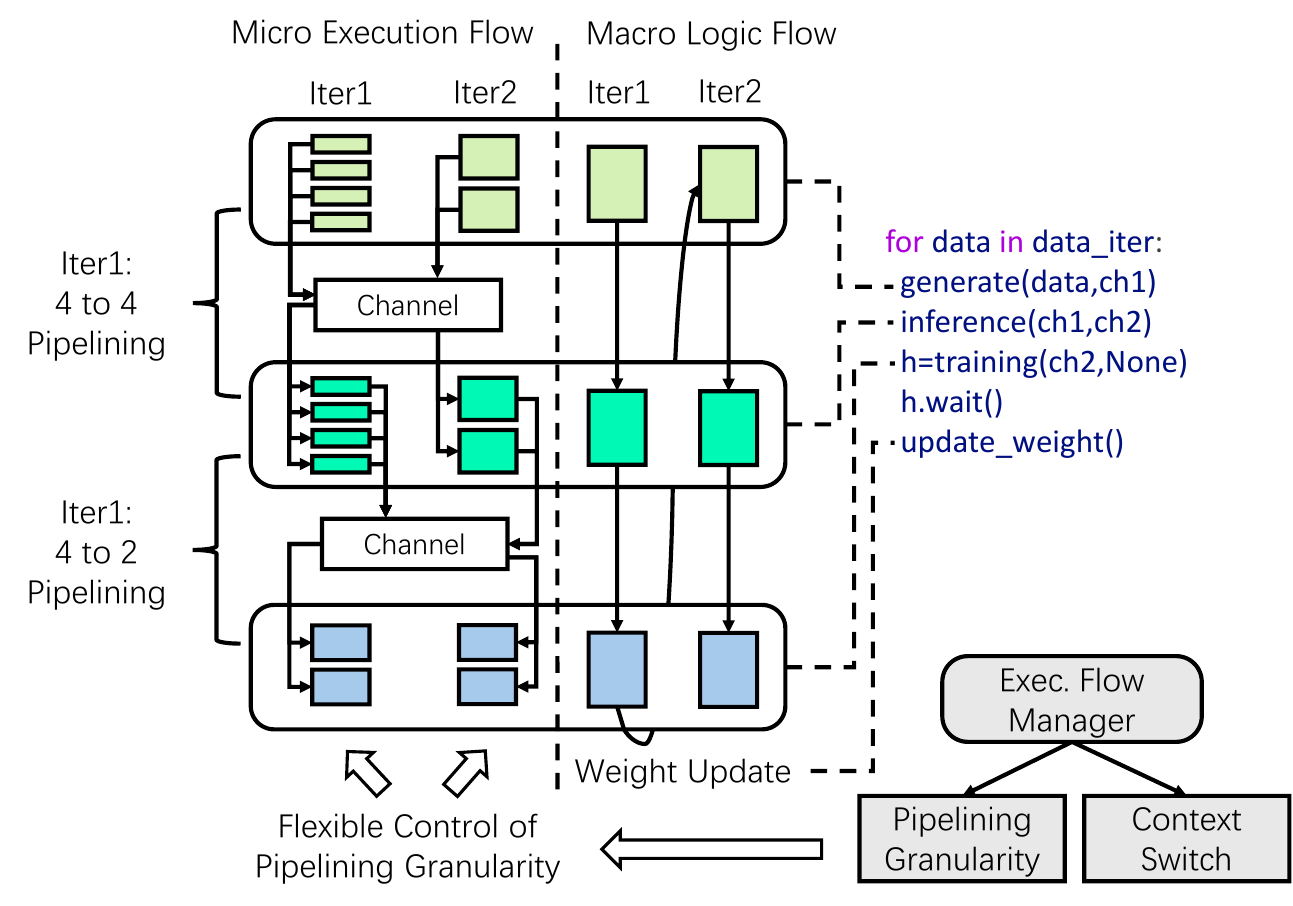}
    \caption{The M2Flow execution logic.}
    \label{fig:m2flow}
    \vspace{-10pt}
\end{figure}

\begin{figure*}[t]
    \centering
    \includegraphics[width=0.8\linewidth]{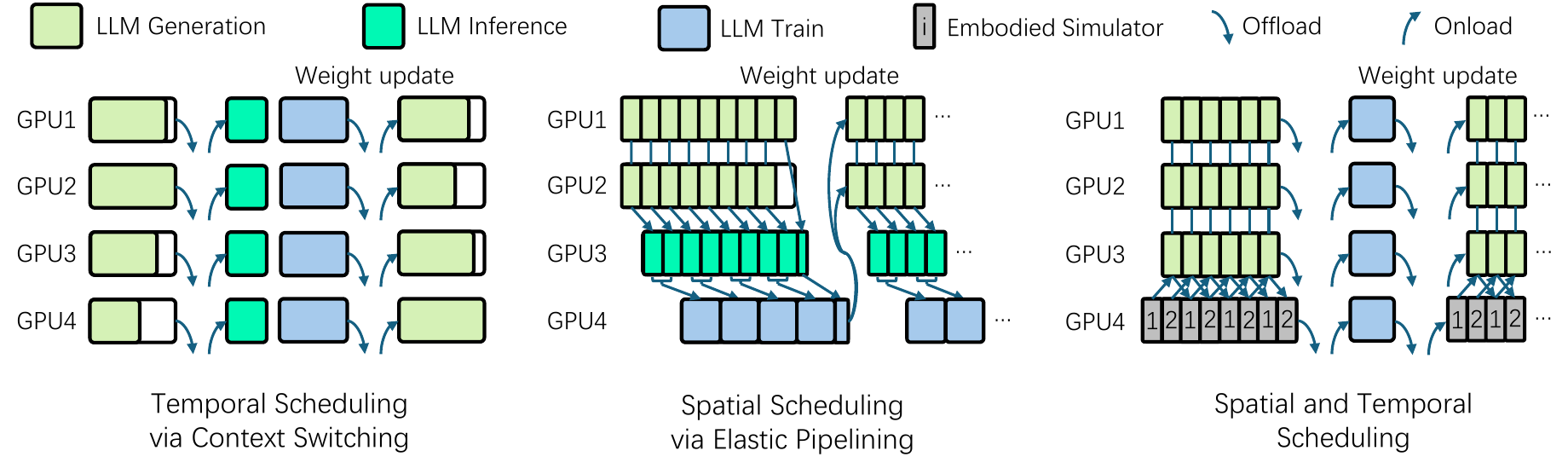}
    \vspace{-10pt}
    \caption{Spatial and temporal scheduling of workers.}
    \label{fig:worker_scheduling}
    \vspace{-15pt}
\end{figure*}

The programming interface shown in \autoref{fig:programinterface_runner} offers a flow-like programming model,
  with which developers describe the high-level, logical control and data flows among workers.
Having obtained the logical flows,
  \system follows the M2Flow paradigm to transform the logical flows (i.e., how should the workers run) into the concrete execution flow, i.e., where (spatial) and when (temporal) should the workers run.
In this section, we focus on two enabling mechanisms of M2Flow transformation and flexible scheduling---elastic pipelining and context switching.
In \S\ref{sec:design_scheduling}, we will describe the scheduling policy that determines the optimal execution flow.

Based on the programmed workflow, 
  the key idea of M2Flow transformation is to control the spatial and temporal scheduling of workers by throttling their data processing granularity and concurrent resource accesses, respectively.

\para{Spatial Scheduling via Elastic Pipelining.}
For spatial scheduling, 
  workers can be executed in a pipelined manner with different number of accelerators/devices.
To maximize pipeline flexibility, \system introduces elastic pipelining to enable workers to flexibly process data at different granularity with the given device resources.
Elastic pipelining builds upon our insight that in RL training and agentic scenarios, 
  most workers follow the SPMD pattern, 
  allowing execution across varying batch sizes. 
For instance, LLM serving engines like SGLang and vLLM can process a single prompt or a list of prompts at a time,
  and inference (i.e., prefill-only computation of a model) similarly supports single-batch or multi-batch execution. 
This flexibility enables the Execution Flow Manager of \system to achieve flexible pipelining of a worker task via dynamic data granularity---output data can be forwarded once a configured size of data batch is ready, 
  allowing downstream workers to start earlier with smaller batches or later with larger batches.
Notably, the scheduling space is further affected by individual worker's internal computation semantics.
For example, training workers operate with both the micro-batch and global-batch concepts---the micro-batch defines forward/backward units, while the global-batch determines when model updates occur. 

\para{Temporal Scheduling via Automatic Context Switching.}
Beyond spatial scheduling, 
  \system also supports natural temporal multiplexing of devices via automatic context switching,
  further expanding the scheduling space.
Context switching enables workers that cannot co-reside in the same accelerators with limited device resources (e.g., GPU memory) to share devices by executing sequentially.
In \system, this is realized via a distributed device lock of the data channel facility, i.e., \texttt{device\_lock} as shown in \autoref{fig:programinterface_worker}. 
This lock serves as the primitive to throttle concurrent resource access by multiple workers on the same devices and have data flow dependencies, 
  i.e., producers and consumers of the same channel.
Before using device resources, 
  a worker must acquire the lock,
  whose state is globally consistent to all workers and can only be changed atomically, 
  thus ensuring exclusive access to the resources.
Upon lock acquiring, 
  the worker's device resources are automatically loaded by calling the \texttt{onload} function of the worker if the resources have been offloaded.
After completing its task, 
  the worker releases the lock and offloads its resources via the \texttt{offload} function to free up device resources for children workers in the workflow.
Unlike conventional lock, the device lock leverages the data channel's data dependency information to define lock acquiring priority, i.e.,
  children workers that depend on the outputs of parent workers can only acquire the lock after the parents have enqueued data and released the lock,
  so as to avoid lock contention and deadlock.
Also, it utilizes device placement information from the Controller to avoid unnecessary resource loading and offloading when workers are placed on different devices.

\para{M2Flow Execution.}
\autoref{fig:m2flow} shows how M2Flow manages execution. 
The workflow is written imperatively, i.e., a for loop iterating over main logic, with three workers, i.e., rollout, inference, and training. 
A rollout task such as \texttt{generate(data, ch1)} processes data and enqueues results to channel \texttt{ch1}. 
The Execution Flow Manager can divide the input data into smaller chunks, allowing workers to process outputs at a smaller data granularity. 
Alternatively, tasks can be coalesced into fewer, larger sub-tasks with larger data chunks to realize different temporal scheduling.
Meanwhile, the device lock enables automatic resource management among workers with data dependencies,
  enabling spatial scheduling via context switching.

With M2Flow, user-defined workflows can be orchestrated across a complete spatial–temporal scheduling space. 
\autoref{fig:worker_scheduling} illustrates several representative execution modes suited to different RL workloads and configurations.

The left part of \autoref{fig:worker_scheduling} shows pure temporal scheduling, where each worker occupies all accelerators. 
Once a worker completes its task, it is swapped out and the next worker is swapped in. 
For the illustrated case, since inference and training share the same model weights, no offloading or reloading is required. 
This mode is particularly useful when a worker must use all devices to be runnable, such as a large model's training. 
However, it suffers from GPU idleness due to long-tail effects, e.g., the longest response length in the rollout stage determines the overall completion time of the stage.

The middle part illustrates spatial scheduling, where workers are assigned to separate GPUs. 
Because workers depend on one another, pipelining is applied to mitigate idle time. 
Achieving efficient pipelining requires balancing resources across workers so that their execution times align.

Finally, RL workflows are often too complex for purely temporal or spatial scheduling to remain efficient. 
M2Flow therefore supports hybrid scheduling, as shown on the right of \autoref{fig:worker_scheduling}. 
Some workers are distributed across GPUs with pipelined execution, but once the workers' stage completes, 
  they can be swapped out and replaced by successors to continue the workflow.

\begin{algorithm}[t]
\footnotesize
    \caption{Worker scheduling policy.}\label{algo:scheduling}
    \SetAlgoLined
    \SetKwData{Left}{left}
    \SetKwData{This}{this}
    \SetKwData{Up}{up}
    \SetKwProg{Fn}{Function}{:}{end}
    \SetKwFunction{Union}{Union}
    \SetKwFunction{TraverseStCuts}{TraverseStCuts}
    \SetKwFunction{TraverseGpuNum}{TraverseGpuNum}
    \SetKwFunction{ConvertCircleToNode}{ConvertCircleToNode}
    \SetKwFunction{FindSchedule}{FindSchedule}
    \SetKwFunction{PipeliningTime}{PipeliningTime}
    \SetKwInOut{Input}{Input}
    \SetKwInOut{Output}{Output}
    \Input{Workflow graph $G$, execution time estimation function $E$ for each component, and number of devices $N$.}
    \Output{A worker schedule $S_{best}$ and its estimated time $T_{best}$.}
    \BlankLine
    $D_{table}\leftarrow$$\{\}$; \tcp{graph map to (time, schedule)}
    $G_{dag}\leftarrow$\ConvertCircleToNode($G$)\;
    $T_{best},S_{best}\leftarrow$\FindSchedule($G_{dag}$, $N$, $D_{table}$)\;
    \BlankLine
    \Fn{\FindSchedule{$G$, $N$, $D_{table}$}}{
        \If{($G$,$N$) in $D_{table}$}{
            \Return $D_{table}$[($G$,$N$)]\;
        }
        \If{$G$ is a node}{
            \Return $E_{node}$, $S_{node}$\;
        }
        $T_{best}\leftarrow$+$inf$; $S_{best}\leftarrow$None\;
        \For{$G_s$, $G_t$ in \TraverseStCuts($G$)}{
            \CommentSty{/* $G_s$ and $G_t$ share the same gpus */}\\
            $T_s,S_s\leftarrow$\FindSchedule($G_s$, $N$, $D_{table}$)\;
            $T_t,S_t\leftarrow$\FindSchedule($G_t$, $N$, $D_{table}$)\;
            $D_{table}$[($G_s$,$N$)]$\leftarrow$($T_s$, $S_s$)\;
            $D_{table}$[($G_t$,$N$)]$\leftarrow$($T_t$, $S_t$)\;
            \If{$T_{best}>T_s+T_t$}{
                $T_{best}\leftarrow$$T_s$+$T_t$; $S_{best}\leftarrow$shared($S_s$, $S_t$)\;
            }
            \CommentSty{/* $G_s$ and $G_t$ use different devices */}\\
            \For{$N_s$, $N_t$ in \TraverseGpuNum($N$)}{
                \CommentSty{/* $N_s$+$N_t$ equals $N$ */}\\
                $T_s,S_s\leftarrow$\FindSchedule($G_s$, $N_s$, $D_{table}$)\;
                $T_t,S_t\leftarrow$\FindSchedule($G_t$, $N_t$, $D_{table}$)\;
                $D_{table}$[($G_s$,$N_s$)]$\leftarrow$($T_s$, $S_s$)\;
                $D_{table}$[($G_t$,$N_t$)]$\leftarrow$($T_t$, $S_t$)\;
                \If{$T_{best}>$\PipeliningTime($T_s,T_t$)}{
                    $T_{best}\leftarrow$\PipeliningTime($T_s,T_t$)\; $S_{best}\leftarrow$pipeline($S_s$, $S_t$)\;
                }
            }
        }
        \Return $T_{best}$, $S_{best}$\;
    }
\end{algorithm}

\subsection{Scheduling Policy}
\label{sec:design_scheduling}
\system offers a large scheduling space through flexible orchestration of workers,
    thus finding the most efficient execution mode is challenging.
To this end, \system introduces two modules: the profiler and the scheduler.
The profiler is used to measure and estimate the execution characteristics of each component under different numbers of GPUs.
The scheduler then utilizes this information to composite an overall execution plan, containing specific GPU assignments and pipelining configurations.

\para{Profiler.} The profiler measures each component’s execution time and memory usage under different data parallel sizes, because data parallel size tend to have positive (sometimes nearly linear) relation with the component throughput as data is consumed faster with the increasing data parallel size.
For non-model components like simulators, the data parallel size is the number of instances.
For model training and generation components, the profiler requires users to provide model parallel configuration to decide the data parallel size, which can usually be determined by the model size and GPU memory.
With the profiled data, the profiler extrapolates the execution time and memory usage for larger data parallel sizes using polynomial extrapolation, outputting an execution time estimation function $E$ for each component.
This function is then fed to the scheduler for it to estimate each component's execution time and whether it can be fitted into memory given a selected number of GPUs.
Moreover, during profiling, the overall RL workflow is also captured into a workflow graph in a just-in-time manner, by tracing the data flow among workers through the communication primitives.

\para{Scheduler.} The scheduling policy of the scheduler is shown in Algorithm~\ref{algo:scheduling}.
The input of the policy includes the workflow graph $G$, the execution time estimation function $E$ produced by the profiler, and the total number of GPU devices.
Specifically, the scheduling algorithm recursively partitions the workflow graph into two subgraphs, $G_s$ and $G_t$, connected by directed edges known as the \textit{s–t cuts}~\cite{ford2015flows}. For each partition, it evaluates the time cost of both the temporal and spatial scheduling policies. 
In the temporal scheduling, $G_s$ and $G_t$ share the same set of devices (e.g., GPUs)---$G_s$ processes its batch, and upon completion $G_t$ consumes its output. 
In the spatial scheduling, $G_s$ and $G_t$ are assigned to disjoint device sets and executed in a pipelined fashion.
The scheduler uses the profiling results to find the optimal device allocation and data processing granularity for each subgraph.
The algorithm then selects the most performant scheduling, 
  and applies this process recursively until each subgraph reduces to a single node, 
  at which point the node returns its profiled execution time under the assigned placement.

Modeling the execution time of $G_s$ and $G_t$ is the key to finding the optimal scheduling. 
In the temporal scheduling where workers share devices, 
  the cost is the sum of $G_s$ and $G_t$ plus any resource offloading and reloading overhead. 
In the spatial scheduling, the runtime is estimated as
$$T_{\text{critical}}+(M/m-1)\times T_{\text{bottleneck}},$$
  where $T_{\text{critical}}$ is the pipeline warm up and cool down time, 
  $T_{\text{bottleneck}}$ is the runtime of the slowest subgraph, 
  $M$ is the total batch size, 
  and $m$ is the data processing granularity.

Before invoking \texttt{FindSchedule}, the workflow graph is preprocessed to collapse cycles into single nodes. 
When recursion reaches such a node, its computation is evenly partitioned across GPUs. 
This avoids exhaustive partition enumeration while still achieving near-optimal performance.

\subsection{Adaptive Communication}
\label{sec:design_communication}

To support flexible worker orchestration, 
  \system's communication layer needs to realize two key design goals.
\textbf{(1) Flexible.} Any two workers should be able to communicate with each other, regardless of the worker placement and program logic.
\textbf{(2) Adaptive.} Communication primitives should be able to adapt to arbitrary data living in different devices (CPUs and GPUs across nodes), while achieving maximum throughput of the underlying communication links.

However, existing collective communication libraries for GPU and CPU data, e.g., NCCL~\cite{nccl}, Gloo~\cite{gloo}, and MPI~\cite{mpi}, 
  fail to meet our goals as they are mostly built for standard communication patterns across a fixed number of processes in traditional model training and serving scenarios.
In contrast, RL components manifest considerable spatial and temporal dynamicity.
Spatially, components can collocate on the same device, or being distributed across different devices.
Temporally, components can be launched or terminated at arbitrary times.
However, state-of-the-art libraries like NCCL does not support flexible rank scaling and efficient intra-device communications.
Furthermore, the data communicated between components can be complex data structures beyond standard contiguous GPU/CPU data buffer, 
  e.g., a composition of multiple data buffer with varying sizes.
Efficient handling of such data dynamicity is also missing in existing libraries.

\para{Communication Protocol and Primitives.}
To achieve the design goals, 
  we devise worker and data placement-aware communication protocol and primitives,
  which enhances existing CPU/GPU communication libraries in terms of both flexibility and performance in the RL scenario.

At the protocol level,
  \system features transparent connection lifecycle management,
  which avoids manual connection management as in traditional communication libraries,
  and handles dynamic worker placement and scaling automatically.
Specifically,
  upon launch, each worker's placement, IP and port information will be registered into a global worker manager.
Connections among workers are then established with the information lazily when workers invoke communication primitives to reduce connection overhead and enhance scalability.
When a group of workers establish a connection, 
  the connection metadata is maintained both locally by workers and globally by a connection manager.
When a worker is terminated,
  the connection manager will notify all connected workers to teardown the connection and release resources.

At the primitive level,
  like existing libraries, \system offers both synchronous and asynchronous \texttt{send} and \texttt{recv} primitives for point-to-point communication,
  as well as collective communication primitives like \texttt{broadcast}.
Differently,
  \system's primitives automatically exploit the worker and data placement information of the communicating workers to select the most efficient communication backend,
  e.g., NCCL for GPU-GPU communication, zero-copy cudaIPC for intra-GPU communication, and Gloo for CPU communication.
For data dynamicity,
  both the asynchronous and synchronous primitives support arbitrary Python objects as the communication payload,
  which are serialized and deserialized in a structure-aware manner,
  i.e., data buffers are extracted from the objects and communicated directly without serialization/deserialization overhead.
Also, data structure information is piggybacked in the communication metadata to facilitate efficient deserialization at the receiver side.

\para{Load-Balancing Data Channel.}
Atop the above communication primitives, we further build a high-level FIFO queue-like communication facility (termed \textit{data channel}) for producer-consumer worker communication in the workflow.
This enables decoupling of both the control and data flows of producer and consumer workers,
  which is essential for flexible pipelining.
The data channel maintains its data queue in a special channel worker process,
  and can be accessed by any other worker processes by passing the channel handle.
The channel supports both CPU and GPU data,
  and can be configured for offloading GPU data to CPU to reduce GPU memory consumption.
Furthermore,
  the data channel is enhanced with load-balancing capability.
Each item enqueued to the channel can be assigned a weight value,
  which is used to balance the load across multiple consumers dequeuing from the channel.
The consumers can also define custom load-balancing policies,
  which are invoked by the channel upon each dequeue operation to select the desired items from the channel.

\section{Implementation}
\label{sec:implementation}
\system is implemented in 20K lines of Python code.
Among them, 5K lines are for the core worker, controller, and scheduler components.
2K for common workers such as rollout workers based on serving engines like SGLang, vLLM and HuggingFace Transformers, training actors based on Megatron and FSDP, and embodied simulators,
    which can be used out of the box for any future RL models and workflows.
The remaining 13K lines are mostly rich supports for various RL algorithms like PPO and GRPO.
Notably, a typical workflow runner like the LLM reasoning RL workflow implementation is less than 100 lines of code,
    and requires no code changes to be scheduled both temporally and spatially.

Currently,
    \system supports not only traditional LLM-based reasoning RL, but also agentic RL with tooling, and embodied RL involving complex workloads like 3D rendering, physics simulation, and robotic control.
For RL algorithms,
    \system supports popular algorithms like PPO~\cite{ppo}, GRPO~\cite{grpo}, DAPO~\cite{yu2025dapo}, and REINFORCE++~\cite{hu2025reinforce++}, as well as some of their off-policy asynchronous versions.
For models,
    we have implemented support for language models like Qwen~\cite{qwen2025qwen25technicalreport}, multi-modal models like VLM~\cite{bai2025qwen2}, and embodied models like VLA~\cite{kim24openvla, kim2025fine} and Pi0~\cite{openpi}.
The rich RL workflow, algorithm and model support not only accelerates the development of new RL workflows with \system,
    but also demonstrates its generality, versatility and extensibility in practice.

\para{Cluster Management and Device Allocation.}
\system leverages Ray to realize cluster management, launching worker processes on remote nodes, and dispatching worker function executions.
\system does not rely on Ray for device allocation to workers,
    because Ray only supports rigid packed-style (i.e., consecutive devices) or spread-style (i.e., spread-across-node first) resource allocation~\cite{ray},
    which are not flexible enough for today's complex and dynamic RL workflows.
Instead, \system offers a flexible device allocation strategy,
    which allows each worker process to be allocated with any device or devices of any node across the cluster,
    by simply specifying the target devices' global IDs.
Also, hardware devices beyond accelerators such as robot arms are also abstracted and managed as schedulable devices in the same manner as accelerators.
This enables \system to scale on all kinds of hardware that foundation models can interact with.

\para{Performance Profiling.}
Performance profiling support is not only crucial for developers to understand system bottlenecks,
    but also serves as the key to our scheduling policy.
Thus, \system provides worker-group-level timer for every public function invoked remotely by the workflow runner.
The timer automatically captures the execution time of a worker function,
    whose value can be retrieved via the asynchronous handle returned by the corresponding worker group function;
    the values of all processes in the worker group will be reduced to a single value via a specified reduction method (e.g., mean, max, min).
Beyond this, developers can also create custom timers for more fine-grained profiling of any code region,
    and retrieve the timer values similarly.

\section{Evaluation}

We extensively evaluate \system across math-reasoning and embodied RL workloads, covering four different models of different sizes (i.e., Qwen2.5~\cite{qwen2025qwen25technicalreport}, Qwen3-MoE~\cite{yang2025qwen3}, OpenVLA~\cite{kim24openvla}, OpenVLA-OFT~\cite{kim2025fine}), two RL algorithms (i.e., GRPO~\cite{grpo}, PPO~\cite{ppo}), and multiple cluster scales. Overall, our key findings include:
\squishlist
\item \system consistently outperforms state-of-the-art RL systems  veRL~\cite{verl} and Slime~\cite{slime} by 1.07$\times$$\sim$1.70$\times$ on a variety of math-reasoning RL settings. The results also show that different RL settings favor different execution modes.
\item \system demonstrates higher training throughput on embodied RL tasks in both ManiSkill~\cite{taomaniskill3} and LIBERO~\cite{liu2023libero}.
On LIBERO, \system achieves a 1.05$\times$$\sim$2.43$\times$ speedup over SimpleVLA-RL~\cite{simplevla};
on ManiSkill, its hybrid mode yields up to 1.87$\times$ improvement over other strategies.
\item The scheduling policy identifies the best execution mode within $7\times 10^{-4}\sim5.98s$ on the clusters of 8 to 1024 GPUs.
\item The models trained with \system achieve SOTA or better benchmark scores after RL fine-tuning.
\squishend


\begin{figure}[t]
    \centering
    \includegraphics[width=0.95\linewidth]{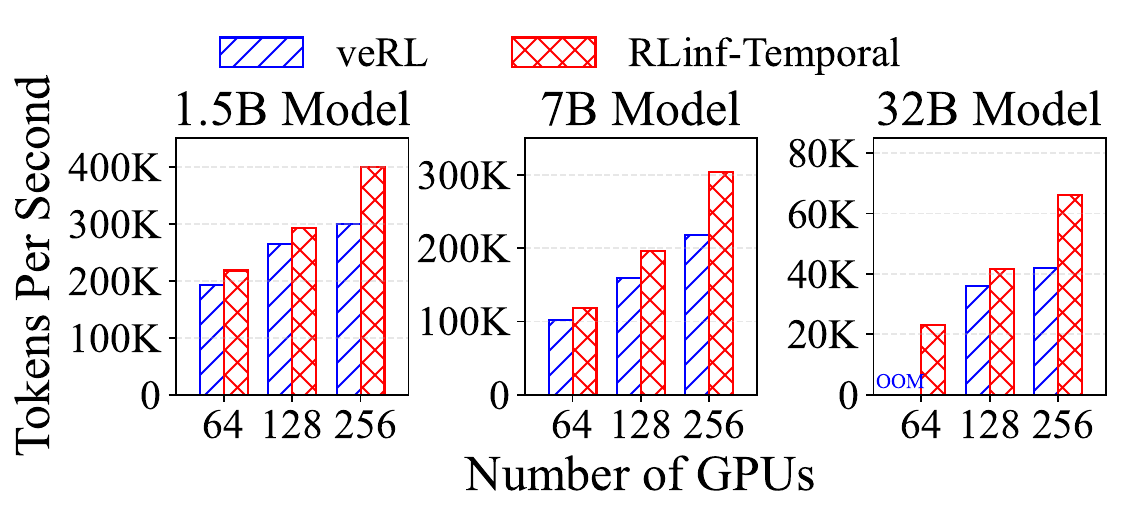}
    \vspace{-10pt}
    \caption{GRPO training throughput of Qwen2.5 on \system and veRL under different cluster scales and model sizes.}
    \label{fig:eval-grpo-shared-verl-all}
    \vspace{-15pt}
\end{figure}

\subsection{End-to-End Experiments}


The end-to-end experiments run on a 32-node cluster, where each node is equipped with 8 NVIDIA H100-80GB GPUs, 2 Intel Xeon Platinum 8558 CPUs (2.1 GHz, 48 cores), and 2 TB of memory. Intra-node communication uses NVLink, while inter-node communication uses 8 Mellanox ConnectX-7 RDMA NICs per node, each providing 400 Gbps bandwidth with RoCEv2.

\subsubsection{Reasoning RL Training}
\para{Experimental setup.}
We evaluate using Qwen2.5 Models~\cite{qwen2025qwen25technicalreport} distilled by DeepSeek-R1,covering sizes from 1.5B to 32B parameters and Qwen3-30B-A3B~\cite{yang2025qwen3} which is an MoE model, and using AReaL-boba-Data dataset~\cite{arealboba}, which integrates multiple standard datasets including DeepScaleR, Open-Reasoner-Zero, Light-R1, etc.
We compare \system with two state-of-the-art open-source RL systems, i.e., veRL v0.5~\cite{verl} and Slime v0.1~\cite{slime}. To ensure fairness, all systems use SGLang~\cite{sglang} for rollouts and Megatron-LM~\cite{megatron} for training, with the same parallelism setting (e.g., tensor parallelism configuration).
Training speed is reported in tokens/sec, defined as the total number of prompt and response tokens in a global batch divided by the iteration time. All performance results are averaged over 10 training iterations after warm-up.

\begin{figure}[t]
    \centering
    \includegraphics[width=0.9\linewidth]{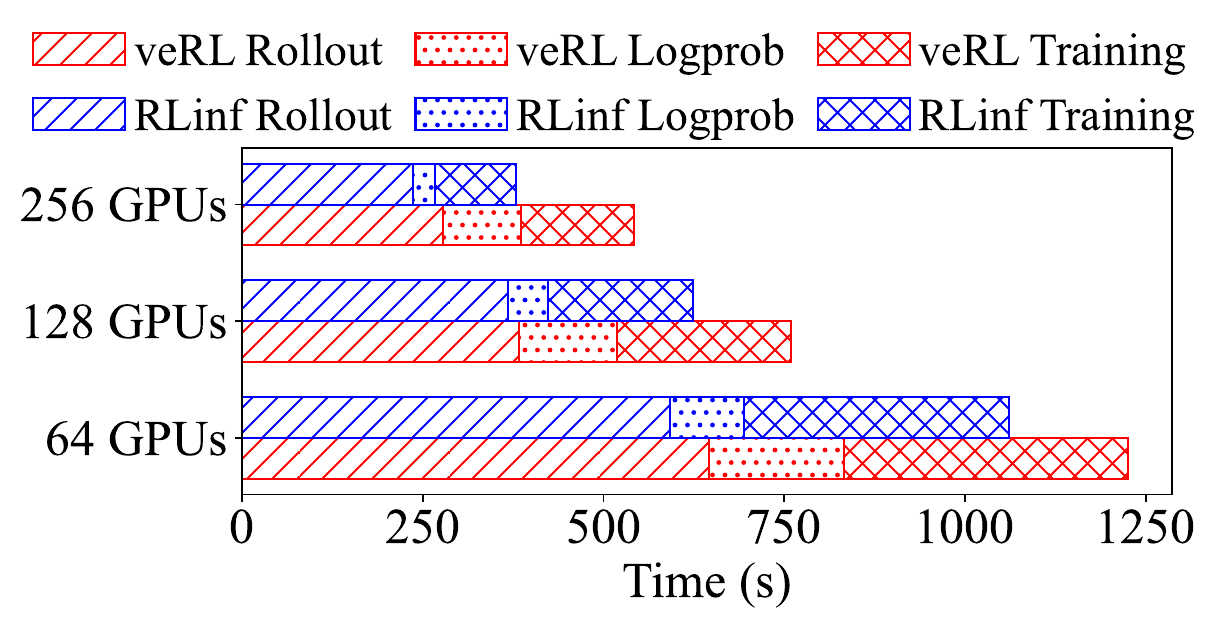}
    \vspace{-10pt}
    \caption{Latency breakdown of Qwen2.5 7B model training.}
    \label{fig:eval-latency_breakdown_7B}
    \vspace{-10pt}
\end{figure}

\para{Qwen2.5 with GRPO.}
We evaluate the Qwen2.5 1.5B, 7B, and 32B dense models on 64, 128, and 256 GPUs, respectively, using a rollout batch size of 512 and maximum sequence length 28672. As shown in \autoref{fig:eval-grpo-shared-verl-all}, \system in temporal mode consistently outperforms veRL across all model scales, achieving 1.10$\times$ to 1.58$\times$ speedups. Since \system’s temporal mode is similar to veRL’s design, the gains primarily stem from faster rollouts enabled by larger KV-cache allocations (\system has better GPU memory management) and reduced synchronization overhead between upstream and downstream stages of the middle inference stage, as illustrated in \autoref{fig:eval-latency_breakdown_7B}.

\autoref{fig:eval-grpo-shared-verl-all} also shows that veRL scales poorly as GPU number increases, largely because inference becomes a growing bottleneck, with its proportion of total execution time rising from 15.2\% to 19.9\% when scaling from 64 to 256 GPUs. In addition, veRL’s unoptimized rollout engine leads to excessive peak memory usage, forcing smaller KV-cache allocations and causing rollout time to decrease only sublinearly.

We also tested spatial mode, but it underperforms veRL by 44.3-68.6\% for the 7B mode, because rollout, inference, and training run on disjoint smaller sets of GPUs, slowing both rollout and training. With the long sequence length (i.e., 28672), the training stage also waits longer for the first batch of rollouts to be generated. In addition, the rollout batch size is relatively smaller, reducing the overlapping time.



\begin{figure}[t]
    \centering
    \includegraphics[width=0.95\linewidth]{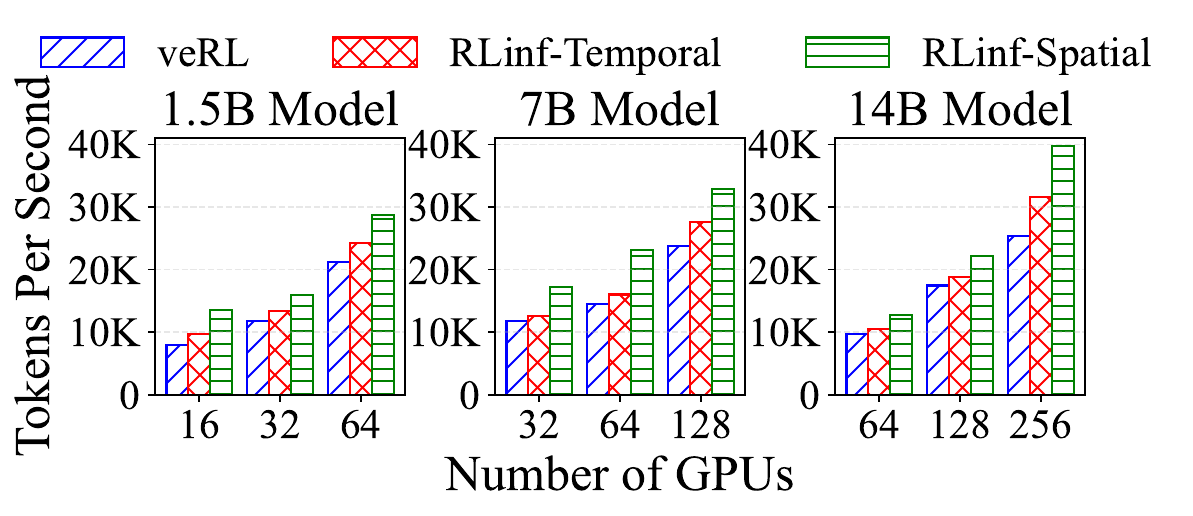}
    \vspace{-10pt}
    \caption{PPO training throughput of Qwen2.5 on \system and veRL under different cluster scales and model sizes.}
    \label{fig:eval-rlinf-vs-verl-ppo}
    \vspace{-15pt}
\end{figure}


\begin{figure*}[t]
    \begin{minipage}[t]{0.35\textwidth}
        \centering
        \includegraphics[width=\linewidth]{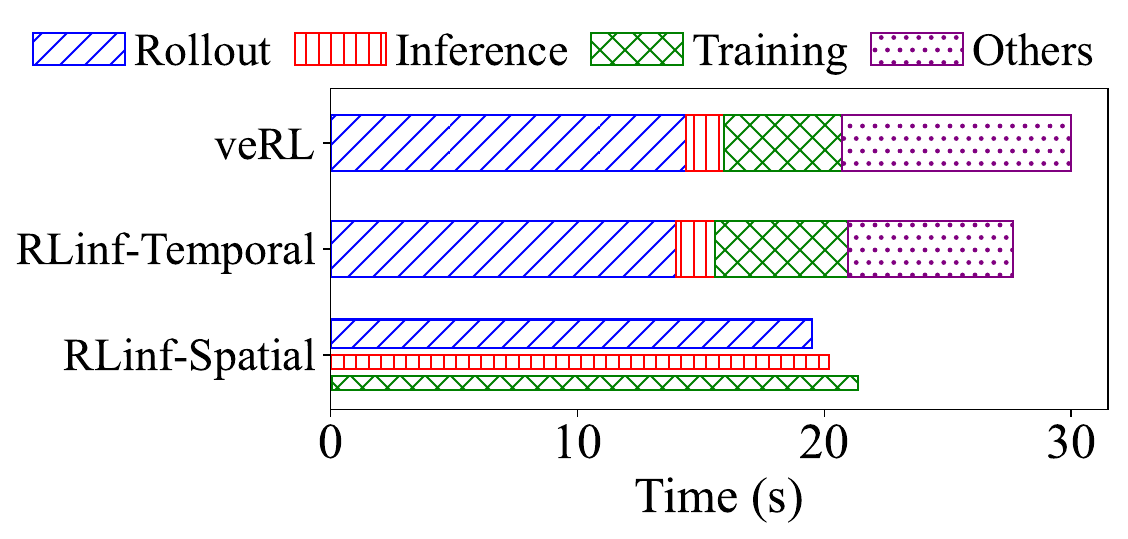}
        \vspace{-20pt}
        \caption{Latency breakdown of Qwen2.5 7B with PPO on 32 GPUs. The width of bar presents the number of GPUs.}
        \label{fig:eval-rlinf-vs-verl-breakdown}
    \end{minipage}
    \hfill
    \begin{minipage}[t]{0.25\textwidth}
        \centering
        \includegraphics[width=\linewidth]{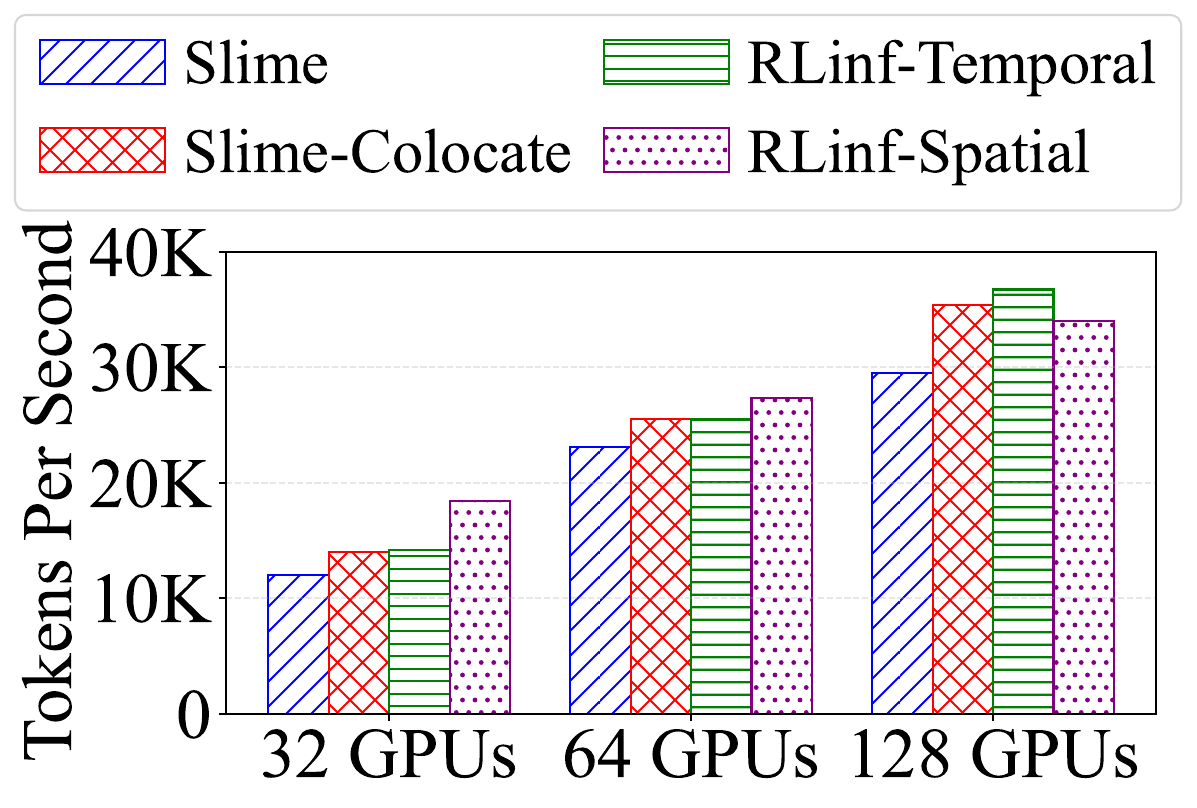}
        \vspace{-20pt}
        \caption{RL training throughput of Qwen3-30B-A3B on \system and Slime under different cluster scales.}
        \label{fig:eval-moe-grpo-shared-verl-all}
    \end{minipage}
    \hfill
        \begin{minipage}[t]{0.35\textwidth}
        \centering
        \includegraphics[width=\linewidth]{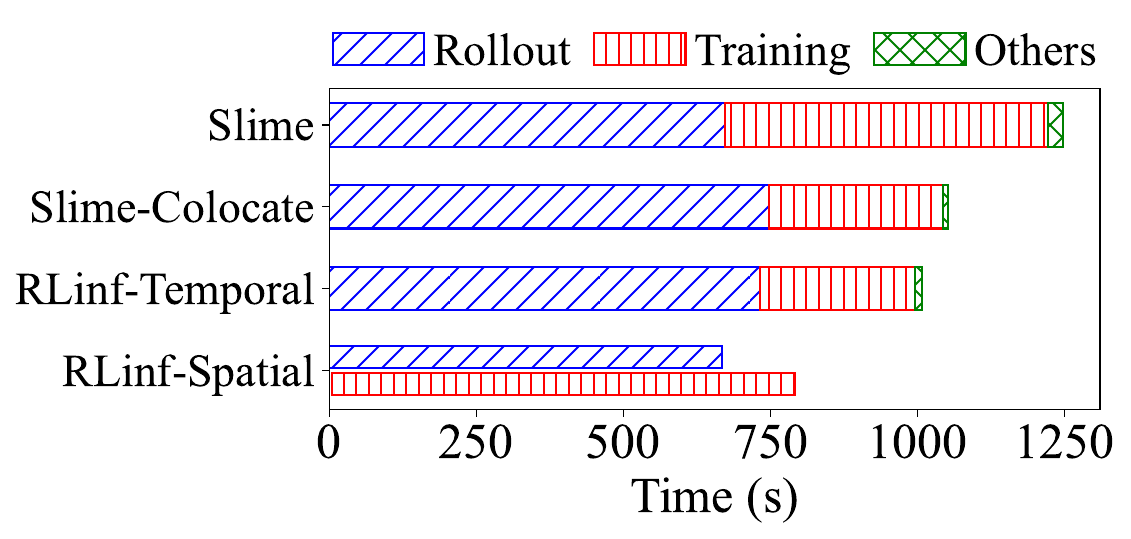}
        \vspace{-20pt}
        \caption{Latency breakdown of Qwen3-30B-A3B with GRPO on 32 GPUs. The width of bar is proportional to the number of GPUs.}
        \label{fig:eval-moe-rlinf-vs-verl-breakdown}
    \end{minipage}
    \hfill
    \vspace{-10pt}

\end{figure*}

\para{Qwen2.5 with PPO.}
We train PPO with Qwen2.5 as both the actor and critic, using model sizes of 1.5B, 7B, and 14B, scaling from 16 to 256 GPUs. We use a rule-based reward, a maximum sequence length of 12288, and a rollout batch size ranging from 256 to 1024 depending on GPU count. \system-Spatial allocates GPUs in a 4:1:1:1:1 ratio for rollout, actor inference, actor training, critic inference, and critic training. \autoref{fig:eval-rlinf-vs-verl-ppo} presents the results. For the 1.5B model, \system-Spatial outperforms veRL by 69.6\%, 35.0\%, and 35.6\% on 16, 32, and 64 GPUs, respectively. In contrast to the GRPO experiment, \system-Spatial surpasses \system-Temporal by 39.8\%, 19.4\%, and 18.6\%.

For the 7B model, \system-Spatial outperforms both veRL and \system-Temporal by 38.7–60.7\% and 19.0–44.8\%, respectively. As shown in \autoref{fig:eval-rlinf-vs-verl-breakdown}, \system-Temporal exceeds veRL mainly due to speedups in ``Others'', which includes context-switch overhead (offload/onload), parameter resharding, advantage computation, and inter-stage synchronization. \system-Spatial performs faster than \system-Temporal because rollout, inference, and training overlap effectively. Although rollout in spatial mode is 39.3\% slower than in temporal mode, inference and training finish quickly once rollout completes. The 14B model shows similar trends with smaller gains, i.e., \system-Spatial is 27.2–56.5\% and 17.7–25.7\% faster than veRL and \system-Temporal, respectively.



\para{Qwen3-30B-A3B with GRPO.}
We evaluate the MoE model Qwen3-30B-A3B on 32, 64, and 128 GPUs with a rollout batch size of 1536 and sequence length 20480. We disable logprob recomputation, a common configuration in RL, which removes the inference stage. Since Slime is optimized for MoE RL training, we compare against two variants: Slime (spatial mode without pipelining, as Slime does not support pipelining) and Slime-Colocate (temporal mode). As shown in \autoref{fig:eval-moe-grpo-shared-verl-all}, Slime is the slowest because rollout and training run on disjoint GPU sets with no execution overlap. For 32 and 64 GPUs, \system-Temporal performs similarly to Slime-Colocate, while \system-Spatial (1:1 rollout-to-training GPUs) is 31.2\% and 7.2\% faster than Slime-Colocate, respectively.

\autoref{fig:eval-moe-rlinf-vs-verl-breakdown} presents the performance breakdown. Spatial mode achieves shorter rollout time because temporal mode suffers from memory contention between rollout and training, forcing much smaller parallelism (i.e., \textit{max\_running\_requests} of 128 vs.\ 256 in SGLang). In addition, \system-Spatial achieves good rollout–training overlap.
On 128 GPUs, \system-Temporal outperforms Slime-Colocate by 3.7\% due to better scaling from decoupled system design; both stages run slightly faster. \system-Spatial underperforms because rollout and training do not overlap well, where training continues for 80 seconds after rollout completes.


\subsubsection{Embodied RL Training}
\begin{figure}
    \centering
    \includegraphics[width=0.95\linewidth]{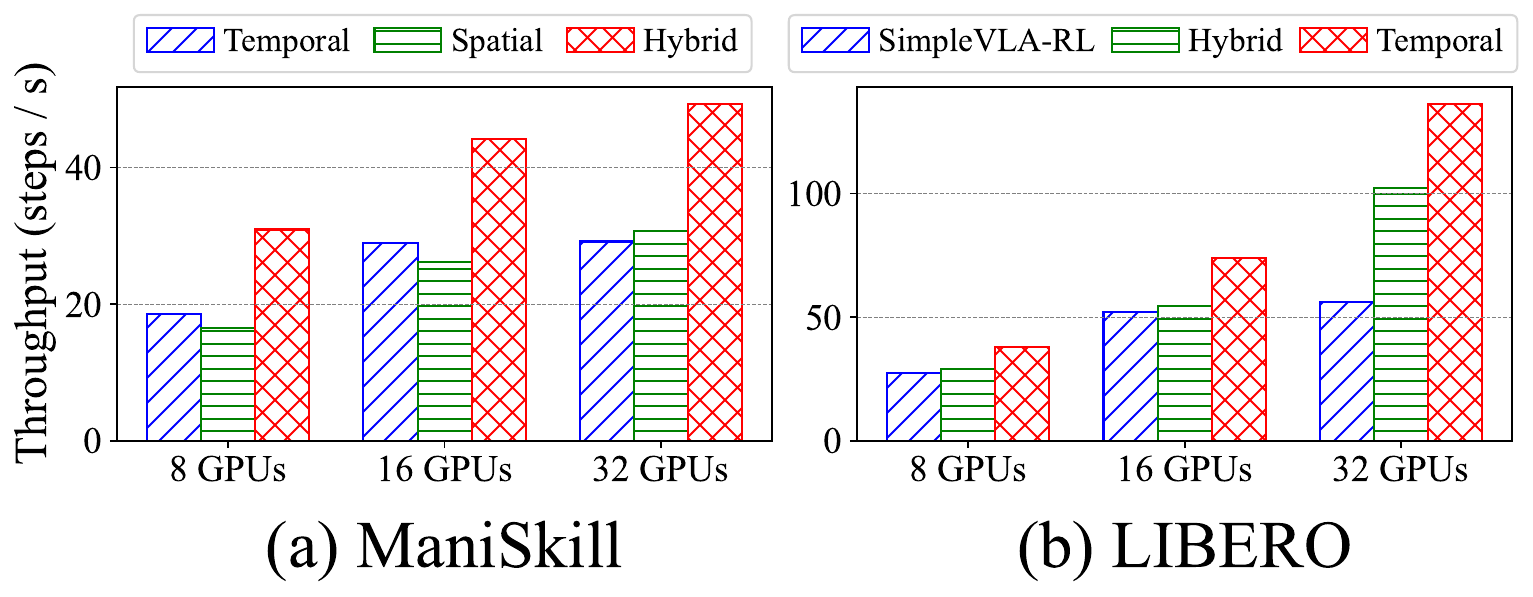}
    \vspace{-10pt}
    \caption{End-to-end throughput of \system and SimpleVLA-RL under different cluster scales.}
    \label{fig:eval-emb-all}
\end{figure}

\para{Experimental setup.}
We evaluate on OpenVLA~\cite{kim24openvla} and OpenVLA-OFT~\cite{kim2025fine}, two vision-language-action models that are supervised-finetuned using RL4VLA~\cite{liu2025what} and SimpleVLA-RL~\cite{simplevla}, respectively. We train OpenVLA on ManiSkill~\cite{taomaniskill3} and OpenVLA-OFT on LIBERO~\cite{liu2023libero}, two widely used embodied environments that emulate physical tasks such as pick-and-place. On LIBERO, we compare \system with SimpleVLA-RL (commit d001d)~\cite{simplevla}, which is built on veRL. On ManiSkill, no distributed RL baseline exists, so we compare different execution modes of \system. Training speed is reported in steps/sec, computed as the total number of environment steps divided by the iteration time.

\para{ManiSkill environment.}
We train OpenVLA on the ``PutCarrotOnPlateInScene-v2'' task~\cite{liu2025what} using 256 parallel ManiSkill environments, each stepping for 80 steps per iteration. \autoref{fig:eval-emb-all}(a) reports end-to-end throughput across three execution modes in \system on 8, 16, and 32 GPUs. \system-Hybrid achieves 52.2–69.1\% higher throughput than \system-Temporal and 60.7–87.2\% higher than \system-Spatial.
For ManiSkill environment, the per-step environment time remains nearly constant as parallelism increases (\autoref{fig:time_vs_envs}) and scaling parallel environments is primarily limited by GPU memory. Thus, dedicating GPUs to environment execution is advantageous. As rollout (i.e., environment and generation) and training all require GPU memory, temporally sharing GPUs between rollout and training is more effective. Thus, \system-Hybrid performs much better than the other two.
In \system-Temporal, environments and generation must coexist in memory, constraining both the number of environments and the batch size, ultimately reducing GPU utilization.
\system-Spatial splits GPUs across the components, forcing training to retain a sufficient number of GPUs to be runnable, which leaves too few GPUs for rollout and significantly slows it down.


\begin{figure}[t]
    \centering
    \includegraphics[width=1\linewidth]{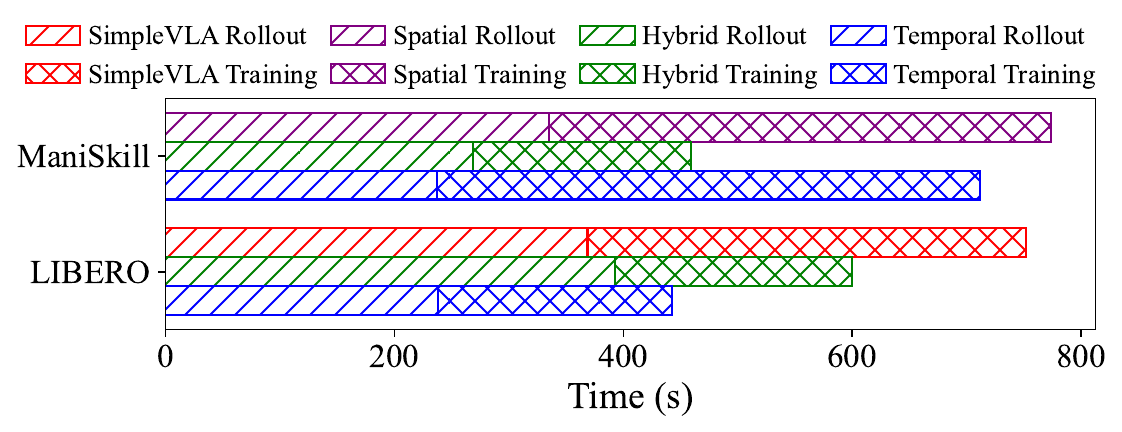}
    \vspace{-20pt}
    \caption{Latency breakdown of ManiSkill and LIBERO.}
    \label{fig:eval-breakdown-libero}
\end{figure}

\para{LIBERO environment.}
We train OpenVLA-OFT on the public benchmark task groups provided by LIBERO, using 512 parallel environments, each running 64 steps per iteration. The results in \autoref{fig:eval-emb-all}(b) show that \system-Temporal is 37.8\%, 42.6\%, 143.4\% faster than SimpleVLA-RL on 8, 16, 32 GPUs respectively. As shown in \autoref{fig:eval-breakdown-libero}, both rollout and training are faster in \system-Temporal than in SimpleVLA-RL due to (1) the elimination of redundant environment initialization during rollout, and (2) the use of a single forward pass to compute both action and log probability, with only a modest memory increase. \system-Temporal also scales better thanks to its decoupled system design and more efficient implementation. In contrast to ManiSkill, \system-Hybrid performs worse than \system-Temporal because LIBERO environment is CPU-intensive; allocating environments on a subset of GPUs limits the utilization of CPU cores, making its rollout stage even slower than that of SimpleVLA-RL.

\begin{figure}[t]
    \centering
    \includegraphics[width=0.95\linewidth]{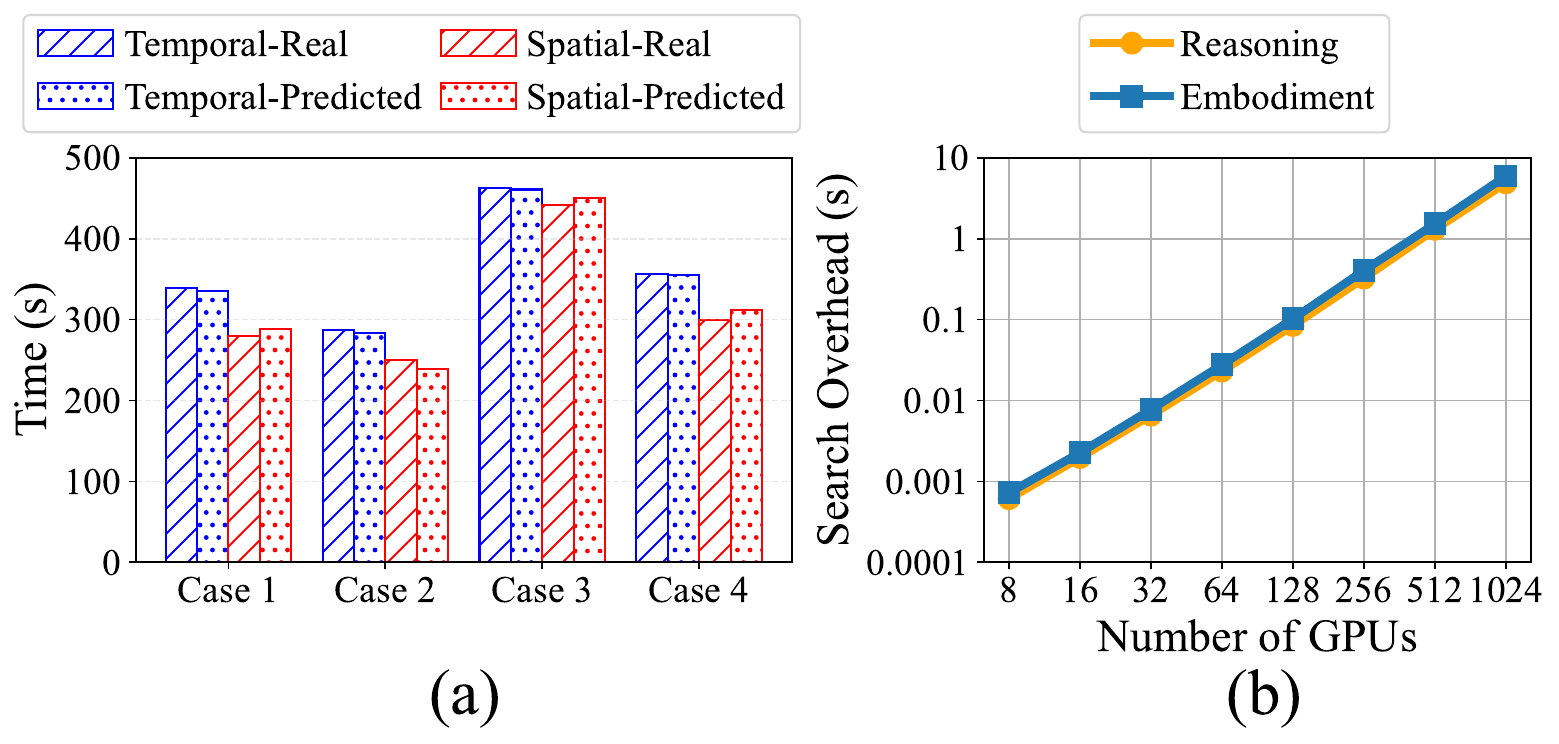}
    \vspace{-10pt}
    \caption{(a) The real and predicted latency in different cases. (b) The search overhead under different cluster scales.}
    \label{fig:eval-static-scheduling}
\end{figure}

\subsection{Effectiveness of Search Policy}
We evaluate the search policy along two dimensions: (a) the accuracy of end-to-end throughput estimates derived from profiling data, and (b) the search speed as the number of GPUs increases.

\para{Estimation accuracy.}
\autoref{fig:eval-static-scheduling}(a) compares real and predicted iteration times across four Qwen2.5–GRPO settings. Cases 1 and 2 use the 1.5B model on 128 GPUs with temperatures 0.6 and 1.0, while Cases 3 and 4 use the 7B model on 64 GPUs with the same temperature settings. Temporal mode achieves estimation errors below 2\% because its end-to-end time is well approximated by summing profiled worker times. Spatial mode with pipelining shows slightly higher errors (<5\%), mainly due to response-length variability that causes pipeline imbalance. These errors are small enough that they do not change the ranking of feasible execution modes, allowing the search policy to consistently select the most efficient mode for the above end-to-end experiments.

\para{Search speed.}
Search complexity is dominated by the number of workflow-graph nodes and available GPUs. RL workflows typically have fewer than ten nodes, allowing our dynamic-programming search to finish quickly. For a graph with three nodes (e.g., Qwen2.5-GRPO or ManiSkill) and GPU counts ranging from 8 to 1024 (\autoref{fig:eval-static-scheduling}(b)), search time grows exponentially with GPU count but remains under five seconds, demonstrating the efficiency of our algorithm.

\subsection{Model Performance}
We comprehensively evaluate the performance of the models produced by the reasoning and the embodied RL training, respectively, to demonstrate the effectiveness of \system.



\begin{table}[t]
\centering
\footnotesize
\caption{Evaluation scores of \system-math-1.5B/7B and open-source models. GPQA represents GPQA-diamond.}
\begin{tabular}{lcccc}
\hline
\textbf{Model} & \textbf{AIME24} & \textbf{AIME25} & \textbf{GPQA} & \textbf{Avg.} \\
\hline
\multicolumn{5}{c}{\textbf{1.5B models}} \\
\hline
Base model & 28.33 & 24.90 & 27.45 & 26.89 \\
DeepScaleR~\cite{deepscaler2025} & 40.41 & 30.93 & 27.54 & 32.96 \\
AReaL~\cite{areal} & 44.42 & 34.27 & 33.81 & 37.50 \\
FastCuRL~\cite{fastcurl} & 43.65 & 32.49 & 35.00 & 37.05 \\
\system & \textbf{48.44} & \textbf{35.63} & \textbf{38.46} & \textbf{40.84} \\
\hline
\multicolumn{5}{c}{\textbf{7B models}} \\
\hline
Base model & 54.90 & 40.20 & 45.48 & 46.86 \\
AReaL~\cite{areal} & 61.66 & 49.38 & 46.93 & 52.66 \\
Skywork~\cite{skywork-or1-2025} & 66.87 & 52.49 & 44.43 & 54.60 \\
Polaris~\cite{Polaris2025} & \textbf{68.55} & 51.24 & 43.88 & 54.56 \\
AceMath~\cite{acemath2024} & 67.30 & \textbf{55.00} & 45.57 & 55.96 \\
\system & 68.33 & 52.19 & \textbf{48.18} & \textbf{56.23} \\
\hline
\end{tabular}
\label{tab:eval-scores-reasoning}
\end{table}

\begin{table}[t]
\centering
\footnotesize
\caption{OpenVLA model success rate results on ManiSkill3.}
\label{tab:maniskill-eval}
\begin{tabular}{lcccc}
\hline
\textbf{Model}      & \textbf{Vision} & \textbf{Semantic} & \textbf{Position} & \textbf{Avg.} \\
\hline
RL4VLA~\cite{liu2025what}  & 80.47\% & 75.00\% & 81.77\% & 79.15\% \\
\system      & \textbf{82.03\%} & \textbf{78.35\%} & \textbf{85.42\%} & \textbf{81.93\%} \\
\hline
\end{tabular}
\vspace{-10pt}
\end{table}


\para{Reasoning RL Training.}
We applied GRPO training, a built-in RL algorithm in \system, on two base models, i.e.,  DeepSeek-R1-Distill-Qwen-1.5B and DeepSeek-R1-Distill-Qwen-7B, to improve their math reasoning ability. As shown in \autoref{tab:eval-scores-reasoning}, the models trained with \system achieve the best average performance compared to the baseline systems. \system 1.5B model outperforms the baselines across all the three benchmarks (i.e., AIME 24~\cite{aime24}, AIME 25~\cite{aime25}, GPQA-diamond~\cite{rein2024gpqa}), improving by up to 20 points over its base model. \system 7B model achieves the highest performance on GPQA-diamond.

\begin{table}[t]
\centering
\caption{OpenVLA-OFT success rate after RL on LIBERO.}
\label{tab:libero-eval}
\resizebox{1.0\columnwidth}{!}{
\begin{tabular}{lccccc}
\hline 
\textbf{Model} & \textbf{Spatial} & \textbf{Object} & \textbf{Goal} & \textbf{Long} & \textbf{Avg.} \\
\hline
OpenVLA-OFT(one-traj) ~\cite{simplevla} & 56.45\% & 25.60\% & 45.59\% & 9.68\% & 34.33\% \\

\system & \textbf{98.99\%} & \textbf{98.99\%} & \textbf{98.99\%} & \textbf{94.35\%} & \textbf{97.83\%} \\

\hline
\end{tabular}
}
\vspace{-10pt}
\end{table}

\para{Embodied RL Training.}
We also evaluate the performance of the models after embodied RL training using \system. \autoref{tab:maniskill-eval} and \autoref{tab:libero-eval} present the evaluation results on the ManiSkill and LIBERO tasks by training OpenVLA and OpenVLA-OFT models respectively, using the PPO algorithm. For ManiSkill, the trained OpenVLA achieves higher success rate than the RL4VLA~\cite{liu2025what} baseline system. For LIBERO, we further train the publicly released OpenVLA-OFT model (one-trajectory fine-tuned version) from SimpleVLA-RL~\cite{simplevla}. By comparing its task success rates with those of the model RL trained using \system, we find that \system significantly improves OpenVLA-OFT’s performance on the LIBERO tasks.


\section{Related Works}
\para{RL Training Frameworks.}
RL training frameworks adopt varying system designs to support large-scale alignment tasks~\cite{rlhf,deeprl}, typically falling into either task-colocated or task-separated execution modes.
For task-colocated systems, DeepSpeed Chat~\cite{deepspeedchat} and veRL\cite{verl} put all training phases on shared GPUs for simplified orchestration. 
In contrast, task-separated systems like NeMo-Aligner~\cite{nemo} and OpenRLHF~\cite{openrlhf} divide components across devices to improve modularity and scalability. 
AReal~\cite{areal} further introduces asynchronous model updates algorithm upon task-separated systems to increase training throughput.
Unlike these frameworks, \system provides a more flexible component-to-device placement mechanism, enabling practitioners to explore and deploy more efficient execution configurations tailored to workload characteristics.

\para{Distributed Training Systems for LLM.}
Distributed training frameworks for LLMs~\cite{fsdp, alpa, colossal-ai} and system-level optimizations for memory~\cite{recompute, offload, stalloc} and communication~\cite{centauri, syndicate, p3} have significantly advanced large-model scaling.
Megatron-LM~\cite{megatron} combines tensor, pipeline, and data parallelism to improve scalability, while DeepSpeed~\cite{deepspeed} introduces the ZeRO family of techniques~\cite{zero, zero-offload, zero-infinity} that shard optimizer states, gradients, and activations to reduce GPU memory usage.
Although distributed RL systems face similar multi-device scaling challenges, RL workflows are more dynamic due to interactive data generation and asynchronous updates, making system design and optimization substantially more complex.

\para{Dataflow System.}
Traditional dataflow systems~\cite{spark, dryad, naiad, mapreduce} excel at large-scale data processing through static task graphs and centralized scheduling, making them efficient for batch and streaming workloads with predictable structures.
In contrast, RL training pipelines feature dynamic task graphs and asynchronous components (e.g., data collection, policy updates), requiring more flexible coordination than existing systems.
Modern frameworks such as Ray~\cite{ray} fill this gap with actor-based execution and decentralized control, making them better suited for orchestrating complex RL workflows.

\section{Conclusion}
Reinforcement learning is poised to surpass pretraining as the driving force behind LLM progress, but its workflows are too diverse and dynamic for rigid execution models. \system shows that by decoupling workflow logic from execution through the novel macro-to-micro transformation mechanism, we can unlock both efficiency and programmability. Beyond RL, we see this approach as a blueprint for future AI runtimes: systems that flexibly orchestrate heterogeneous components, e.g., training, inference, simulation, and reasoning, under one unified execution framework. We believe \system marks an early step toward the operating system for AI workloads.




\bibliographystyle{plain}
\bibliography{reference}

\end{document}